\documentclass[10pt,twocolumn,letterpaper]{article}

\usepackage{cvpr,times,graphicx,subfigure,authblk,amsmath,amssymb,booktabs,multirow}
\usepackage{array,color,bm,ifthen,tabu,xcolor,colortbl,dblfloatfix}
\usepackage[sort]{cite}
\usepackage[pagebackref=true,breaklinks=true,letterpaper=true,colorlinks=true,bookmarks=false]{hyperref}

\makeatletter
\renewcommand\paragraph{\@startsection{paragraph}{4}{\z@}%
  {.5em \@plus1ex \@minus.2ex}%
  {-1em}%
  {\normalfont\normalsize\bfseries}}
\makeatother

\newenvironment{packed_enum}{
\begin{enumerate}
  \setlength{\itemsep}{1pt}
  \setlength{\parskip}{0pt}
  \setlength{\parsep}{0pt}
}{\end{enumerate}}

\cvprfinalcopy
\begin{document}

\title{Learning Features by Watching Objects Move\vspace{-2mm}}
\author[1,2,*]{Deepak Pathak}
\author[1]{Ross Girshick}
\author[1]{Piotr Doll\'ar}
\author[2]{Trevor Darrell}
\author[1]{Bharath Hariharan}
\affil[1]{Facebook AI Research (FAIR)}
\affil[2]{University of California, Berkeley}

\maketitle
\renewcommand*{\thefootnote}{\fnsymbol{footnote}}
\setcounter{footnote}{1}
\footnotetext{Work done during an internship at FAIR.}
\renewcommand*{\thefootnote}{\arabic{footnote}}
\setcounter{footnote}{0}

\begin{abstract}
This paper presents a novel yet intuitive approach to unsupervised feature learning.
Inspired by the human visual system, we explore whether low-level motion-based grouping cues can be used to learn an effective visual representation.
Specifically, we use unsupervised motion-based segmentation on videos to obtain segments, which we use as `pseudo ground truth' to train a convolutional network to segment objects from a single frame.
Given the extensive evidence that motion plays a key role in the development of the human visual system, we hope that this straightforward approach to unsupervised learning will be more effective than cleverly designed `pretext' tasks studied in the literature.
Indeed, our extensive experiments show that this is the case.
When used for transfer learning on object detection, our representation significantly outperforms previous unsupervised approaches across multiple settings, especially when training data for the target task is scarce.
\end{abstract}

\section{Introduction}

ConvNet-based image representations are extremely versatile, showing good performance in a variety of recognition tasks~\cite{donahue2013decaf, fastrcnn, BharathCVPR2015, ZhangECCV2014}.
Typically these representations are trained using \emph{supervised learning} on large-scale image classification datasets, such as ImageNet~\cite{imagenet}.
In contrast, animal visual systems do not require careful manual annotation to learn, and instead take advantage of the nearly infinite amount of unlabeled data in their surrounding environments.
Developing models that can learn under these challenging conditions is a fundamental scientific problem, which has led to a flurry of recent work proposing methods that learn visual representations without manual annotation.

\begin{figure}
\centering
\includegraphics[width=0.495\linewidth]{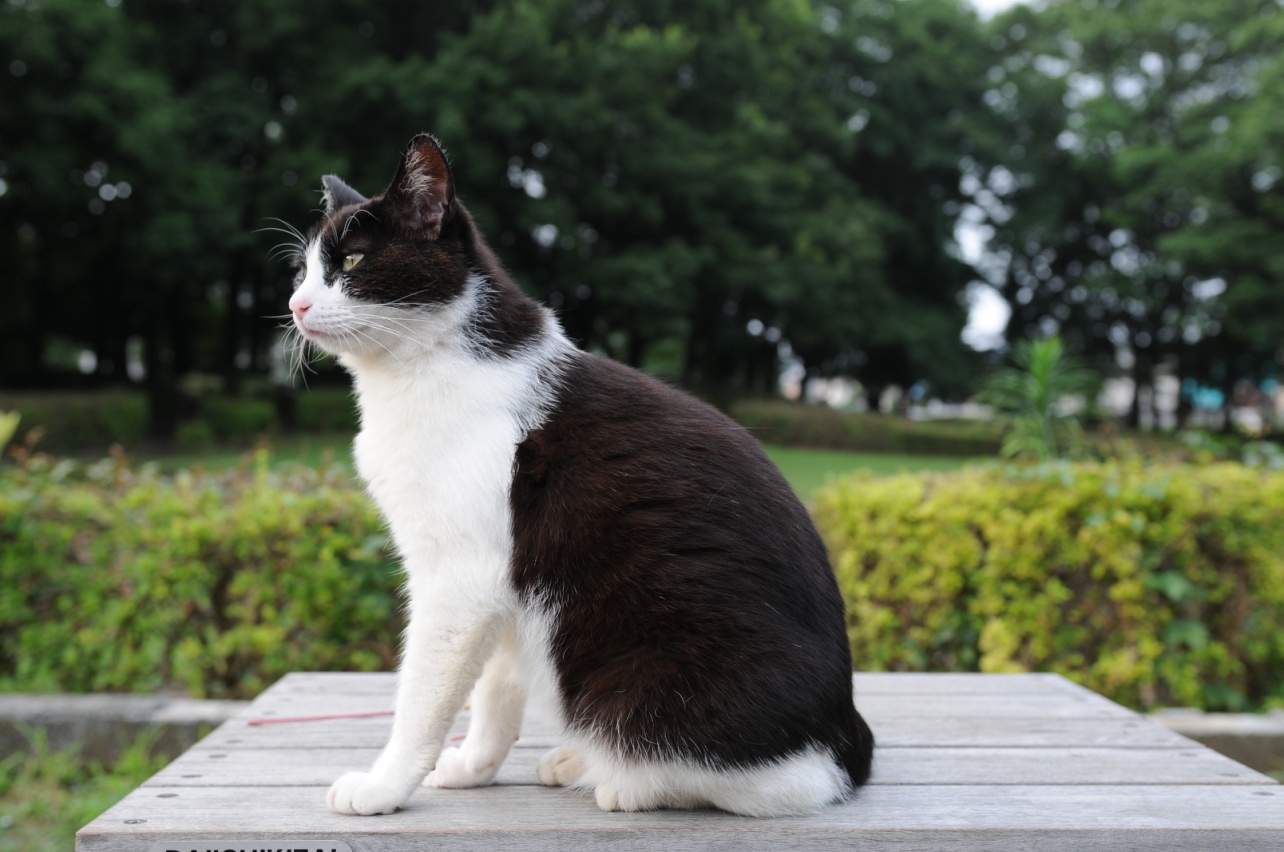}
\includegraphics[width=0.495\linewidth]{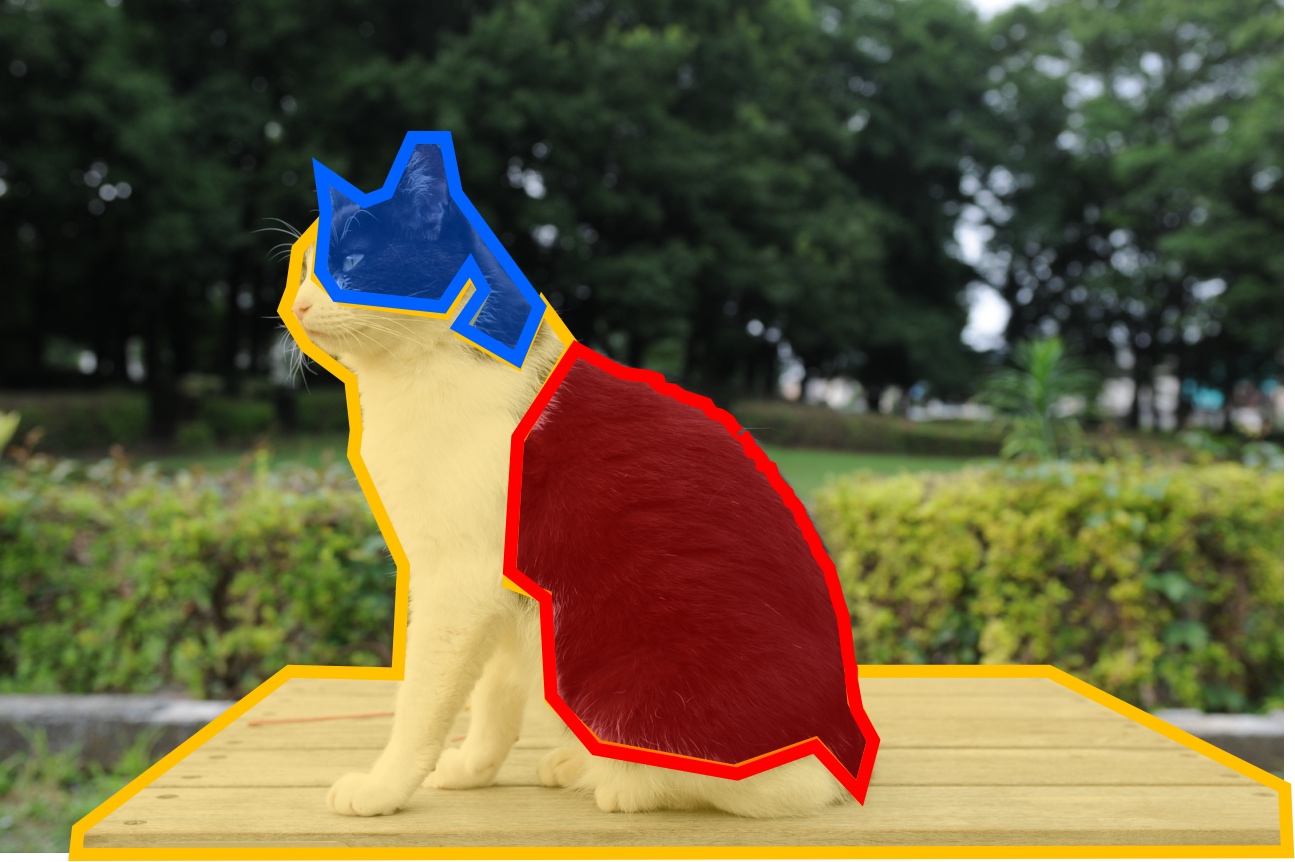}\\
\includegraphics[width=0.495\linewidth]{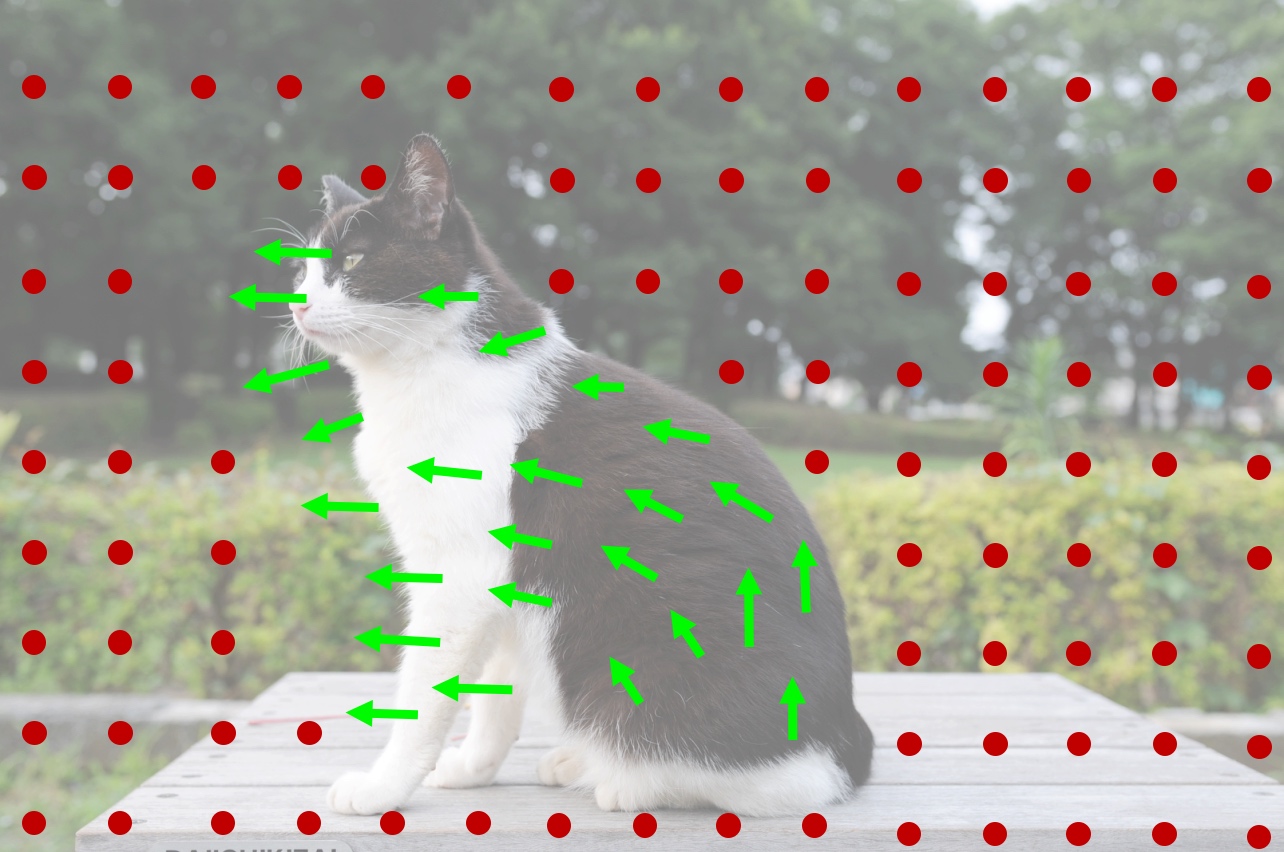}
\includegraphics[width=0.495\linewidth]{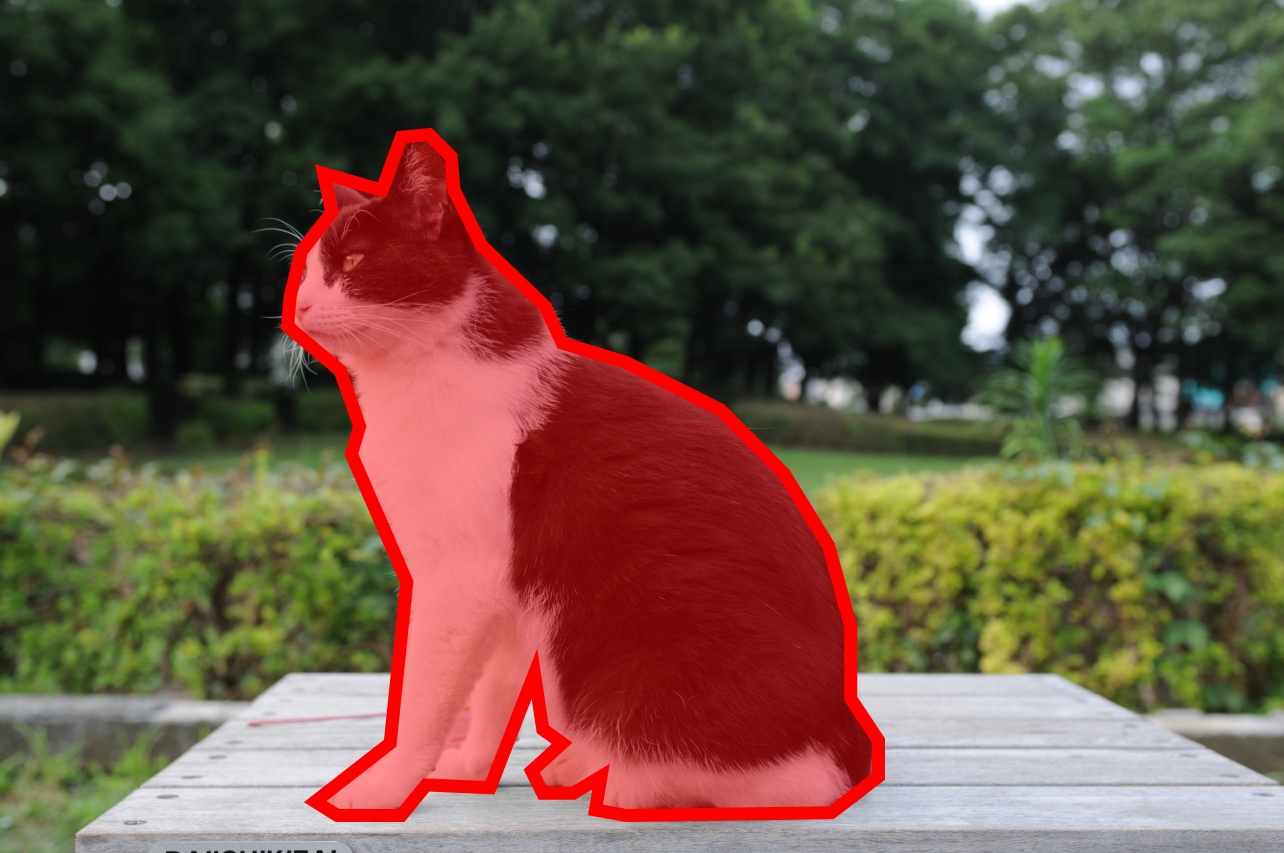}
\caption{Low-level appearance cues lead to incorrect grouping (top right). Motion helps us to correctly group pixels that move together (bottom left) and identify this group as a single object (bottom right). We use unsupervised motion-based grouping to train a ConvNet to segment objects in \emph{static images} and show that the network learns strong features that transfer well to other tasks.}
\label{fig:fig1}
\end{figure}

A recurring theme in these works is the idea of a `pretext task': a task that is not of direct interest, but can be used to obtain a good visual representation as a byproduct of training.
Example pretext tasks include reconstructing the input~\cite{hintonautoenc,bengio2009learning, denoising}, predicting the pixels of the next frame in a video stream~\cite{goroshin2015learning}, metric learning on object track endpoints~\cite{wang2015unsupervised}, temporally ordering shuffled frames from a video~\cite{misra2016unsupervised}, and spatially ordering patches from a static image~\cite{doersch2015unsupervised,NorooziECCV2016}.
The challenge in this line of research lies in cleverly designing a pretext task that causes the ConvNet (or other representation learner) to learn high-level features.

In this paper, we take a different approach that is motivated by human vision studies.
Both infants~\cite{spelke1990principles} and newly sighted congenitally blind people~\cite{ostrovsky2009visual} tend to \emph{oversegment} static objects, but can group things properly when they \emph{move} (Figure~\ref{fig:fig1}).
To do so, they may rely on the Gestalt principle of common fate~\cite{wertheimer1938laws, palmer1999vision}: pixels that move together tend to belong together.
The ability to parse static scenes improves \cite{ostrovsky2009visual} over time, suggesting that while motion-based grouping appears early, static grouping is acquired later, possibly bootstrapped by motion cues.
Moreover, experiments in~\cite{ostrovsky2009visual} show that shortly after gaining sight, human subjects are better able to name objects that tend to be seen in motion compared to objects that tend to be seen at rest.

\begin{figure}
\centering
\includegraphics[width=\linewidth]{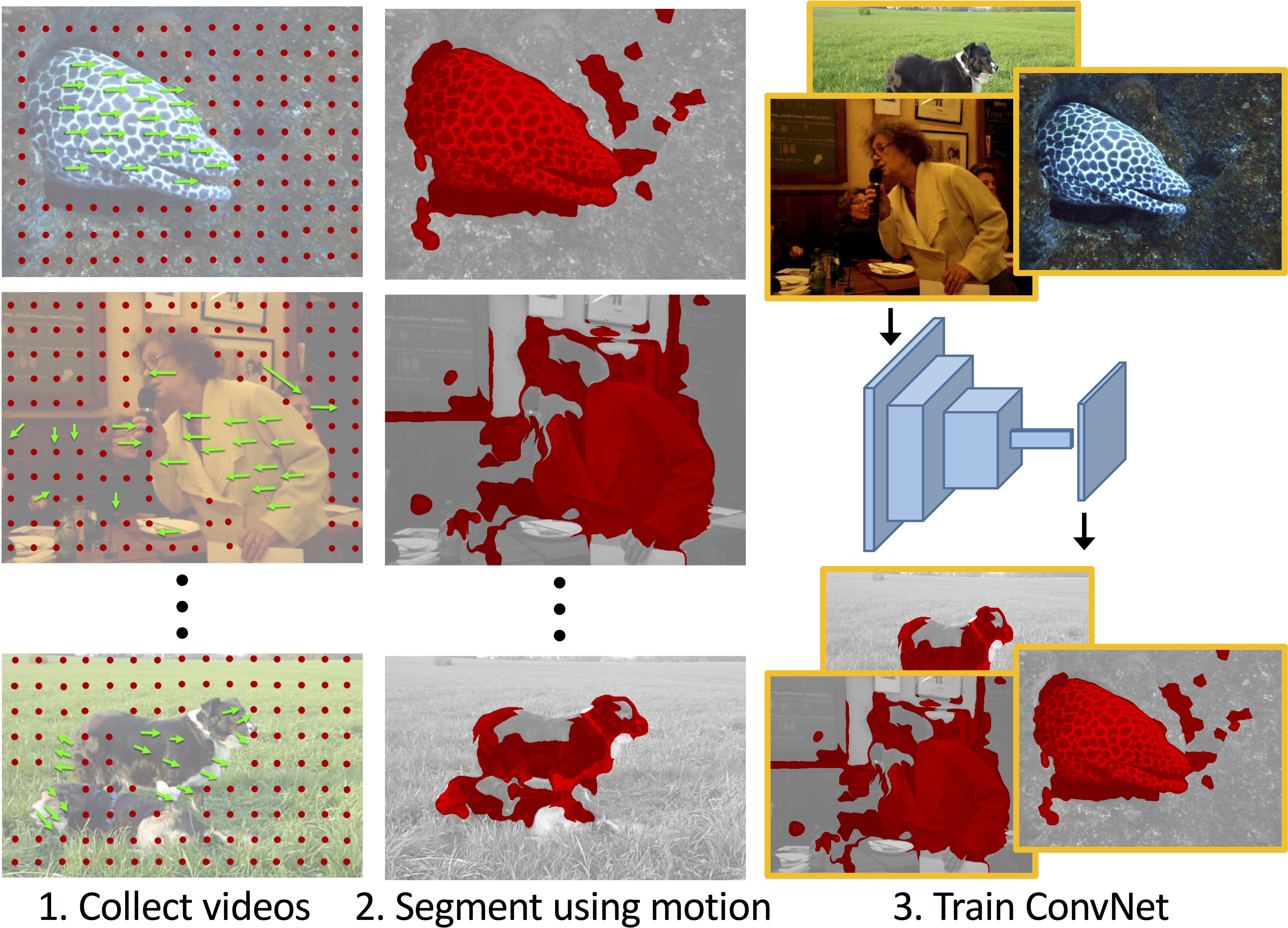}
\caption{Overview of our approach. We use motion cues to segment objects in videos \emph{without any supervision}. We then train a ConvNet to predict these segmentations from \emph{static frames}, \ie without any motion cues. We then transfer the learned representation to other recognition tasks.}
\label{fig:fig2}
\end{figure}

Inspired by these human vision studies, we propose to train ConvNets for the well-established task of object foreground \vs background segmentation, using unsupervised motion segmentation to provide `pseudo ground truth'.
Concretely, to prepare training data we use optical flow to group foreground pixels that move together into a single object.
We then use the resulting segmentation masks as automatically generated targets, and task a ConvNet with predicting these masks from \emph{single, static frames without any motion information} (Figure~\ref{fig:fig2}).
Because pixels with different colors or low-level image statistics can still move together and form a single object, the ConvNet cannot solve this task using a low-level representation.
Instead, it may have to \emph{recognize} objects that tend to move and identify their shape and pose.
Thus, we conjecture that this task forces the ConvNet to learn a high-level representation.

We evaluate our proposal in two settings.
First, we test if a ConvNet can learn a good feature representation when learning to segment from the high-quality, manually labeled segmentations in COCO \cite{lin2014microsoft}, without using the class labels.
Indeed, we show that the resulting feature representation is effective when transferred to PASCAL VOC object detection.
It achieves state-of-the-art performance for representations trained without any semantic category labels, performing within 5 points AP of an ImageNet pretrained model and 10 points higher than the best unsupervised methods.
This justifies our proposed task by showing that \emph{given good ground truth segmentations, a ConvNet trained to segment objects will learn an effective feature representation}.

Our goal, however, is to learn features \emph{without manual supervision}.
Thus in our second setting we train with \emph{automatically generated} `pseudo ground truth' obtained through unsupervised motion segmentation on uncurated videos from the Yahoo Flickr Creative Commons 100 million (YFCC100m)~\cite{thomee2016yfcc100m} dataset.
When transferred to object detection, our representation retains good performance even when most of the ConvNet parameters are frozen, significantly outperforming previous unsupervised learning approaches.
It also allows much better transfer learning when training data for the target task is scarce.
Our representation quality tends to increase logarithmically with the amount of data, suggesting the possibility of outperforming ImageNet pretraining given the countless videos on the web.

\section{Related Work}

Unsupervised learning is a broad area with a large volume of work; Bengio \etal~\cite{bengio2013representation} provide an excellent survey.
Here, we briefly revisit some of the recent work in this area.

\paragraph{Unsupervised learning by generating images.}
Classical unsupervised representation learning approaches, such as autoencoders~\cite{hintonautoenc,bengio2009learning} and denoising autoencoders~\cite{denoising}, attempt to learn feature representations from which the original image can be decoded with a low error.
An alternative to reconstruction-based objectives is to train generative models of images using generative adversarial networks~\cite{goodfellow2014generative}.
These models can be extended to produce good feature representations by training jointly with image encoders~\cite{donahue2016adversarial, dumoulin2016adversarially}.
However, to generate realistic images, these models must pay significant attention to low-level details while potentially ignoring higher-level semantics.

\paragraph{Self-supervision via pretext tasks.}
Instead of producing images, several recent studies have focused on providing alternate forms of supervision (often called `pretext tasks') that do not require manual labeling  and can be algorithmically produced. For instance, Doersch \etal~\cite{doersch2015unsupervised} task a ConvNet with predicting the relative location of two cropped image patches. Noroozi and Favaro~\cite{NorooziECCV2016} extend this by asking a network to arrange shuffled patches cropped from a 3$\times$3 grid. Pathak \etal~\cite{pathakCVPR16context} train a network to perform an image inpainting task. Other pretext tasks include predicting color channels from luminance~\cite{zhang2016colorful, LarssonECCV2016} or vice versa~\cite{zhang2016split}, and predicting sounds from video frames~\cite{owens2016ambient,de1994learning}.
The assumption in these works is that to perform these tasks, the network will need to recognize high-level concepts, such as objects, in order to succeed.
We compare our approach to all of these pretext tasks and show that the proposed natural task of object segmentation leads to a quantitatively better feature representation in many cases.

\paragraph{Learning from motion and action.}
The human visual system does not receive static images; it receives a continuous video stream. The same idea of defining auxiliary pretext tasks can be used in unsupervised learning from videos too. Wang and Gupta~\cite{wang2015unsupervised} train a ConvNet to distinguish between pairs of tracked patches in a single video, and pairs of patches from different videos. Misra~\etal~\cite{misra2016unsupervised} ask a network to arrange shuffled frames of a video into a temporally correct order. Another such pretext task is to make predictions about the next few frames: Goroshin~\etal~\cite{goroshin2015learning} predict pixels of future frames and Walker~\etal~\cite{walker2016uncertain} predict dense future trajectories. However, since nearby frames in a video tend to be visually similar (in color or texture), these approaches might learn low-level image statistics instead of more semantic features. Alternatively, Li \etal~\cite{li2016unsupervised} use motion boundary detection to bootstrap a ConvNet-based contour detector, but find that this does not lead to good feature representations. Our intuitions are similar, but our approach produces semantically strong representations.

Animals and robots can also sense their own motion (proprioception), and a possible task is to predict this signal from the visual input alone~\cite{agrawal2015learning, jayaraman2015learning, GargECCV2016}.
While such cues undoubtedly can be useful, we show that strong representations can be learned even when such cues are unavailable.

\section{Evaluating Feature Representations}
\label{sec:evaluationprotocol}

To measure the quality of a learned feature representation, we need an evaluation that reflects real-world constraints to yield useful conclusions.
Prior work on unsupervised learning has evaluated representations by using them as initializations for fine-tuning a ConvNet for a particular isolated task, such as object detection~\cite{doersch2015unsupervised}.
The intuition is that a good representations should serve as a good starting point for task-specific fine-tuning.
While fine-tuning for each task can be a good solution, it can also be impractical.

For example, a mobile app might want to handle multiple tasks on device, such as image classification, object detection, and segmentation.
But both the app download size and execution time will grow linearly with the number of tasks unless computation is shared.
In such cases it may be desirable to have a general representation that is shared between tasks and task-specific, lightweight classifier `heads'.

Another practical concern arises when the amount of labeled training data is too limited for fine-tuning.
Again, in this scenario it may be desirable to use a fixed general representation with a trained task-specific `head' to avoid overfitting.
Rather than emphasizing any one of these cases, in this paper we aim for a broader understanding by evaluating learned representations under a variety of conditions:
\begin{packed_enum}
\item \textbf{On multiple tasks:} We consider object detection, image classification and semantic segmentation.
\item \textbf{With shared layers:} We fine-tune the pretrained ConvNet weights to different extents, ranging from only the fully connected layers to fine-tuning everything (see~\cite{NorooziECCV2016} for a similar evaluation on ImageNet).
\item \textbf{With limited target task training data:} We reduce the amount of training data available for the target task.
\end{packed_enum}

\section{Learning Features by Learning to Group}
\label{sec:super}

The core intuition behind this paper is that training a ConvNet to \emph{group pixels in static images into objects without any class labels} will cause it to learn a strong, high-level feature representation.
This is because such grouping is difficult from low-level cues alone: objects are typically made of multiple colors and textures and, if occluded, might even consist of spatially disjoint regions.
Therefore, to effectively do this grouping is to implicitly \emph{recognize} the object and understand its location and shape, even if it cannot be \emph{named}.
Thus, if we train a ConvNet for this task, we expect it to learn a representation that aids recognition.

To test this hypothesis, we ran a series of experiments using high-quality manual annotations on static images from COCO~\cite{lin2014microsoft}.
Although \emph{supervised}, these experiments help to evaluate a) how well our method might work under ideal conditions, b) how performance is impacted if the segments are of lower quality, and c) how much data is needed.
We now describe these experiments in detail.

\subsection{Training a ConvNet to Segment Objects}
\label{sec:training_oracle}
We frame the task as follows: given an image patch containing a single object, we want the ConvNet to segment the object, \ie, assign each pixel a label of 1 if it lies on the object and 0 otherwise.
Since an image contains multiple objects, the task is ambiguous if we feed the ConvNet the entire image.
Instead, we sample an object from an image and crop a box around the ground truth segment.
However, given a precise bounding box, it is easy for the ConvNet to cheat: a blob in the center of the box would yield low loss.
To prevent such degenerate solutions, we jitter the box in position and scale.
Note that a similar training setup was used for recent segmentation proposal methods~\cite{DeepMask, SharpMask}.

We use a straightforward ConvNet architecture that takes as input a $w \times w$ image and outputs an $s \times s$ mask.
Our network ends in a fully connected layer with $s^2$ outputs followed by an element-wise sigmoid.
The resulting $s^2$ dimensional vector is reshaped into an $s \times s$ mask.
 We also downsample the ground truth mask to $s \times s$ and sum the cross entropy losses over the $s^2$ locations to train the network.

\subsection{Experiments}
\label{sec:control}
To enable comparisons to prior work on unsupervised learning, we use AlexNet~\cite{krizhevsky2012imagenet} as our ConvNet architecture.
We use $s=56$ and $w = 227$.
We use images and annotations from the trainval set of the COCO  dataset~\cite{lin2014microsoft}, \emph{discarding the class labels} and only using the segmentations.

\begin{figure}[t]
\centering
\includegraphics[width=0.8\linewidth]{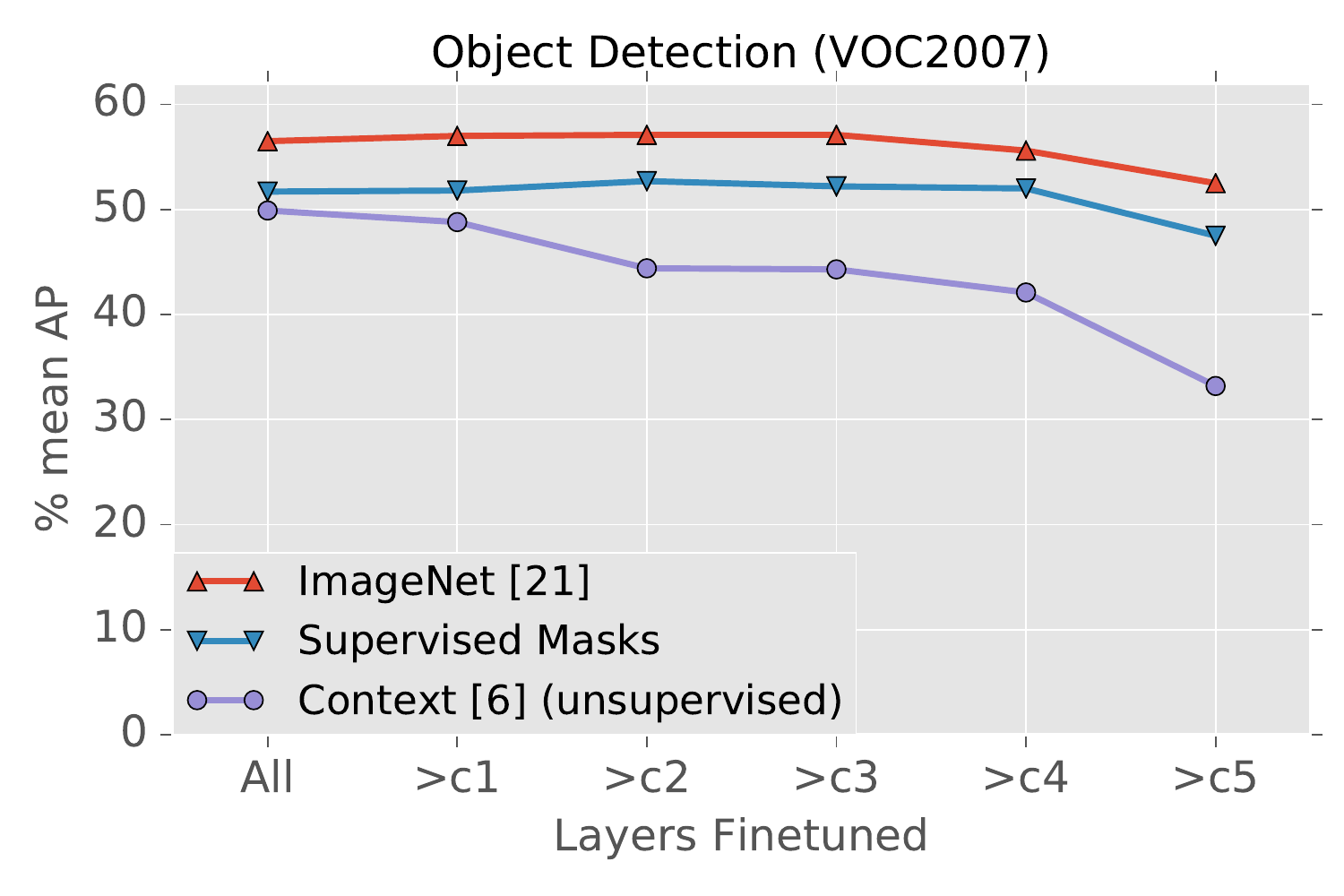}
\caption{Our representation trained on manually-annotated segments from COCO (without class labels) compared to ImageNet pretraining and context prediction (unsupervised)~\cite{doersch2015unsupervised}, evaluated for object detection on PASCAL VOC 2007.
`$>$cX': all layers above convX are fine-tuned; `All': the entire net is fine-tuned.}
\label{fig:supervised}
\end{figure}

\paragraph{Does training for segmentation yield good features?}
Following recent work on unsupervised learning, we perform experiments on the task of object detection on PASCAL VOC 2007 using Fast R-CNN~\cite{fastrcnn}.\footnote{\url{https://github.com/rbgirshick/py-faster-rcnn}}
We use multi-scale training and testing~\cite{fastrcnn}.
In keeping with the motivation described in Section~\ref{sec:evaluationprotocol}, we measure performance with ConvNet layers frozen to different extents.
We compare our representation to a ConvNet trained on image classification on ImageNet, and the representation trained by Doersch \etal~\cite{doersch2015unsupervised}.
The latter is competitive with the state-of-the-art.
(Comparisons to other recent work on unsupervised learning appear later.)
The results are shown in Figure~\ref{fig:supervised}.

We find that our supervised representation outperforms the unsupervised context prediction model across all scenarios by a large margin, which is to be expected.
Notably though, our model maintains a fairly small gap with ImageNet pretraining.
This result is state-of-the-art for a model trained without semantic category labels.
Thus, given high-quality segments, our proposed method can learn a strong representation, which validates our hypothesis.

Figure~\ref{fig:supervised} also shows that the model trained on context prediction degrades rapidly as more layers are frozen.
This drop indicates that the higher layers of the model have become overly specific to the \emph{pretext} task~\cite{yosinski2014transferable}, and may not capture the high-level concepts needed for object recognition.
This is in contrast to the stable performance of the ImageNet trained model even when most of the network is frozen, suggesting the utility of its higher layers for recognition tasks.
We find that this trend is also true for our representation: \emph{it retains good performance even when most of the ConvNet is frozen}, indicating that it has indeed learned high-level semantics in the higher layers.

\paragraph{Can the ConvNet learn from noisy masks?}
We next ask if the quality of the learned representation is impacted by the quality of the ground truth, which is important since the segmentations obtained from unsupervised motion-based grouping will be imperfect.
To simulate noisy segments, we train the representation with degraded masks from COCO.
We consider two ways of creating noisy segments: introducing noise in the boundary and truncating the mask.

\begin{figure}[t]
\includegraphics[width=0.24\linewidth]{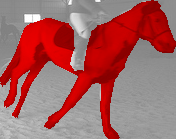}
\includegraphics[width=0.24\linewidth]{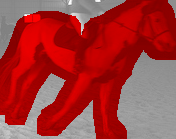}
\includegraphics[width=0.24\linewidth]{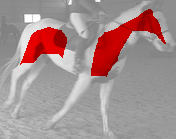}
\includegraphics[width=0.24\linewidth]{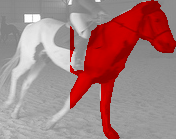}
\caption{We degrade ground truth masks to measure the impact of segmentation quality on the learned representation. From left to right, the original mask, dilated and eroded masks (boundary errors), and a truncated mask (truncation can be on any side).}
\label{fig:degradeegs}
\end{figure}

Noise in the segment boundary simulates the foreground leaking into the background or vice-versa.
To introduce such noise during training, for each cropped ground truth mask, we randomly either erode or dilate the mask using a kernel of fixed size (Figure~\ref{fig:degradeegs}, second and third images).
The boundaries become noisier as the kernel size increases.

Truncation simulates the case when we miss a part of the object, such as when only part of the object moves.
Specifically, for each ground truth mask, we zero out a strip of pixels corresponding to a fixed percentage of the bounding box area from one of the four sides (Figure~\ref{fig:degradeegs}, last image).

\begin{figure}[t]
\centering
\includegraphics[width=0.99\linewidth]{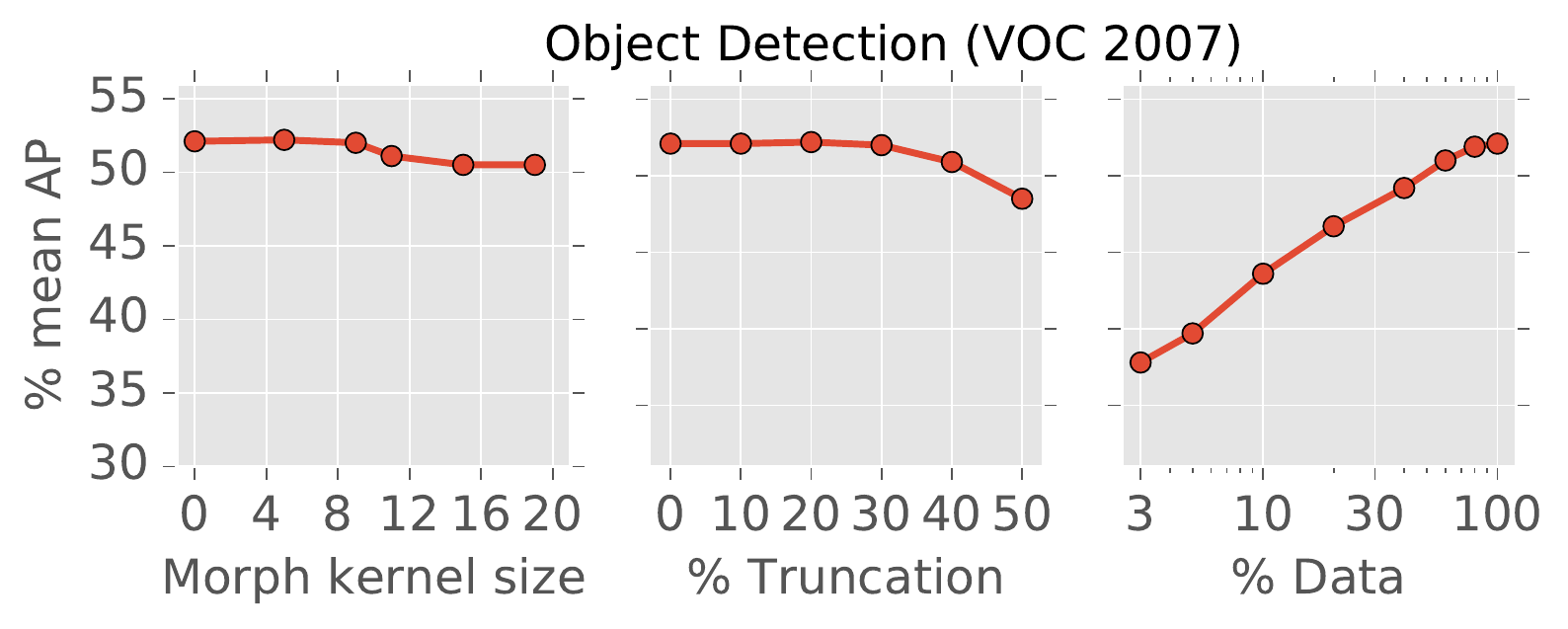}
\caption{VOC object detection accuracy using our \emph{supervised} ConvNet as noise is introduced in mask boundaries, the masks are truncated, or the amount of data is reduced. Surprisingly, the representation maintains quality even with large degradation.}
\label{fig:degrade}
\end{figure}

We evaluate the representation trained with these noisy ground truth segments on object detection using Fast R-CNN with all layers up to and including conv5 frozen (Figure~\ref{fig:degrade}).
We find that the learned representation is surprisingly resilient to both kinds of degradation.
Even with large, systematic truncation (up to 50\%) or large errors in boundaries, the representation maintains its quality.

\paragraph{How much data do we need?}
We vary the amount of data available for training, and evaluate the resulting representation on object detection using Fast-RCNN with all conv layers frozen.
The results are shown in the third plot in Figure~\ref{fig:degrade}.
We find that performance drops significantly as the amount of training data is reduced, suggesting that good representations will need large amounts of data.

In summary, these results suggest that training for segmentation leads to strong features even with imprecise object masks.
However, building a good representation requires significant amounts of training data.
These observations strengthen our case for learning features in an unsupervised manner on large unlabeled datasets.

\section{Learning by Watching Objects Move}

We first describe the motion segmentation algorithm we use to segment videos, and then discuss how we use the segmented frames to train a ConvNet.

\subsection{Unsupervised Motion Segmentation}
The key idea behind motion segmentation is that if there is a single object moving with respect to the background through the entire video, then pixels on the object will move differently from pixels on the background.
Analyzing the optical flow should therefore provide hints about which pixels belong to the foreground.
However, since only a part of the object might move in each frame, this information needs to be aggregated across multiple frames.

\begin{figure}
\includegraphics[width=0.32\linewidth]{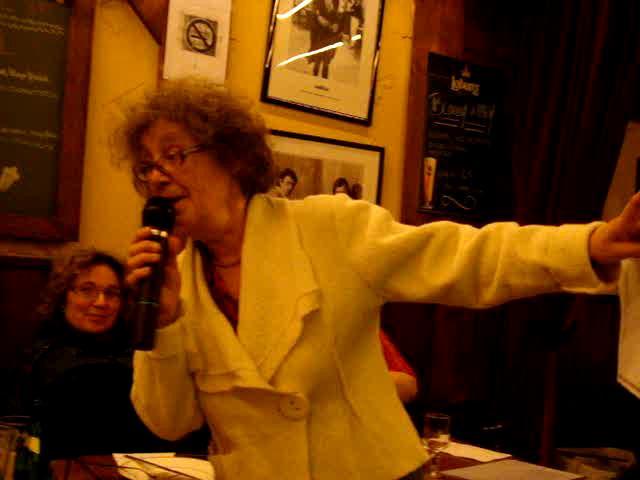}
\includegraphics[width=0.32\linewidth]{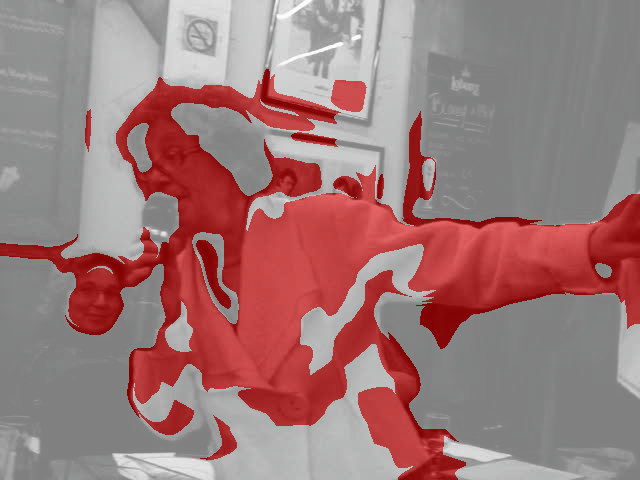}
\includegraphics[width=0.32\linewidth]{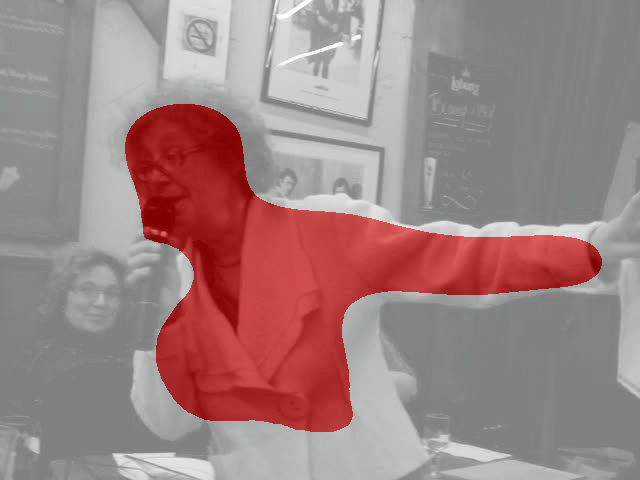}\\
\includegraphics[width=0.32\linewidth]{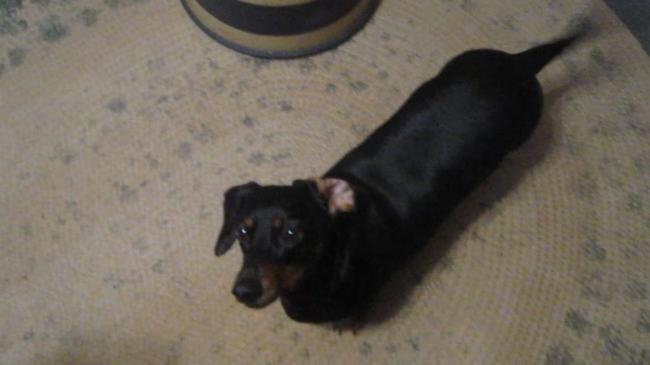}
\includegraphics[width=0.32\linewidth]{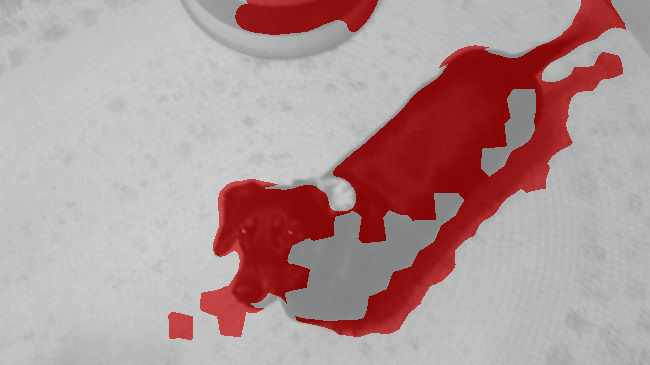}
\includegraphics[width=0.32\linewidth]{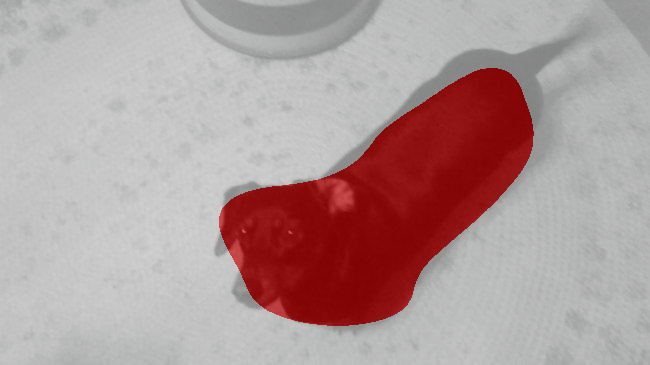}\\
\includegraphics[width=0.32\linewidth]{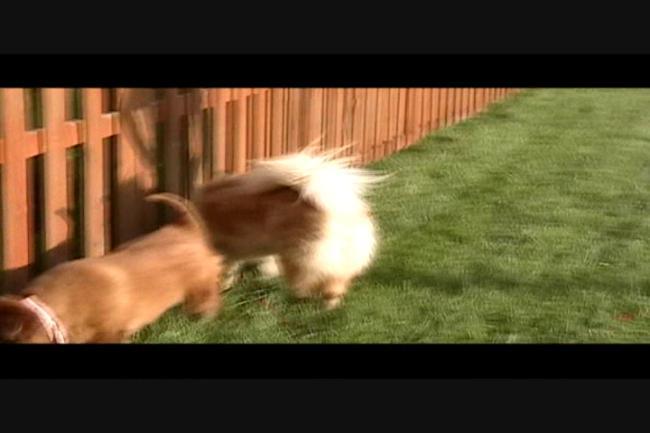}
\includegraphics[width=0.32\linewidth]{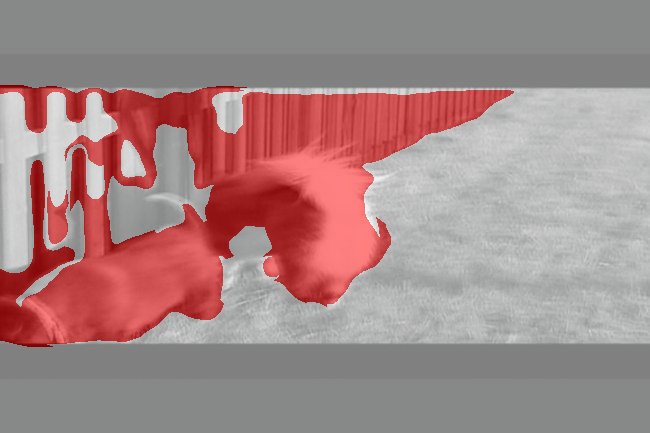}
\includegraphics[width=0.32\linewidth]{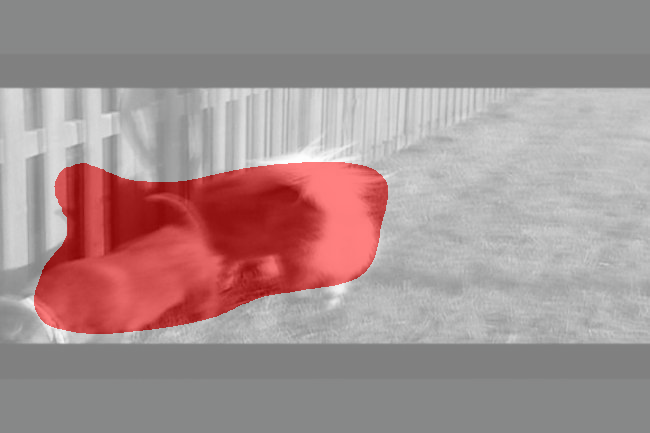}\\
\includegraphics[width=0.32\linewidth]{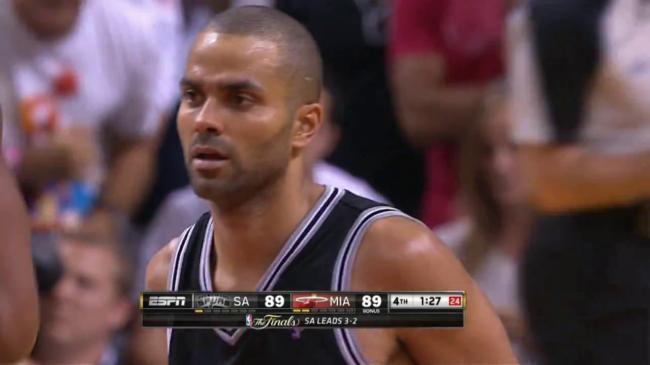}
\includegraphics[width=0.32\linewidth]{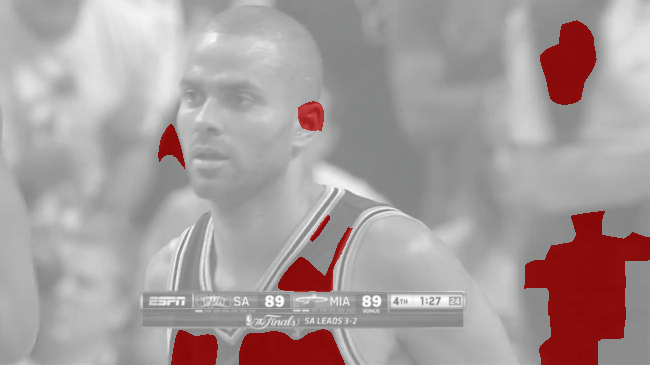}
\includegraphics[width=0.32\linewidth]{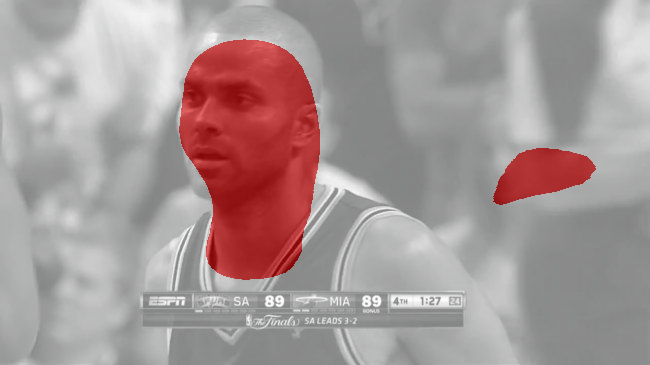}
\caption{From left to right: a video frame, the output of uNLC that we use to train our ConvNet, and the output of our ConvNet. uNLC is able to highlight the moving object even in potentially cluttered scenes, but is often noisy, and sometimes fails (last two rows). Nevertheless, our ConvNet can still learn from this noisy data and produce significantly better and smoother segmentations.}\vspace{-3mm}
\label{fig:motionsegegs}
\end{figure}

We adopt the NLC approach from Faktor and Irani~\cite{faktor2014video}.
While NLC is unsupervised with respect to video segmentation, it utilizes an edge detector that was trained on labeled edge images~\cite{structuredforests}.
In order to have a purely unsupervised method, we replace the trained edge detector in NLC with unsupervised superpixels.
To avoid confusion, we call our implementation of NLC as uNLC.
First uNLC computes a per-frame saliency map based on motion by looking for either pixels that move in a mostly static frame or, if the frame contains significant motion, pixels that move in a direction different from the dominant one.
Per-pixel saliency is then averaged over superpixels~\cite{slic}.
Next, a nearest neighbor graph is computed over the superpixels in the video using location and appearance (color histograms and HOG~\cite{dalal2005histograms}) as features.
Finally, it uses a nearest neighbor voting scheme to propagate the saliency across frames.

\begin{figure}
\includegraphics[width=0.32\linewidth]{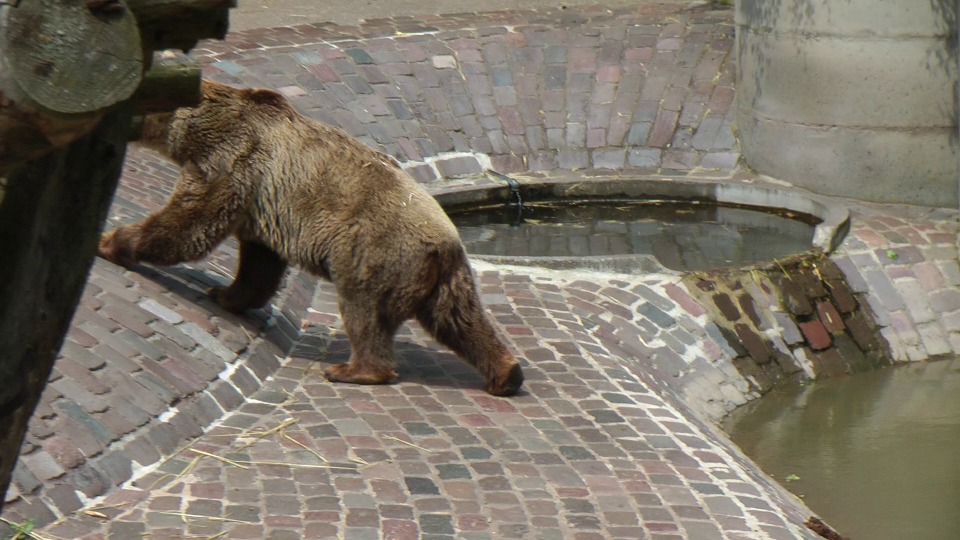}
\includegraphics[width=0.32\linewidth]{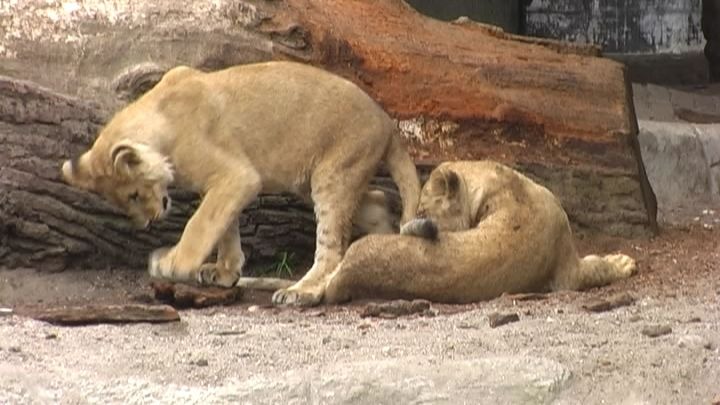}
\includegraphics[width=0.32\linewidth]{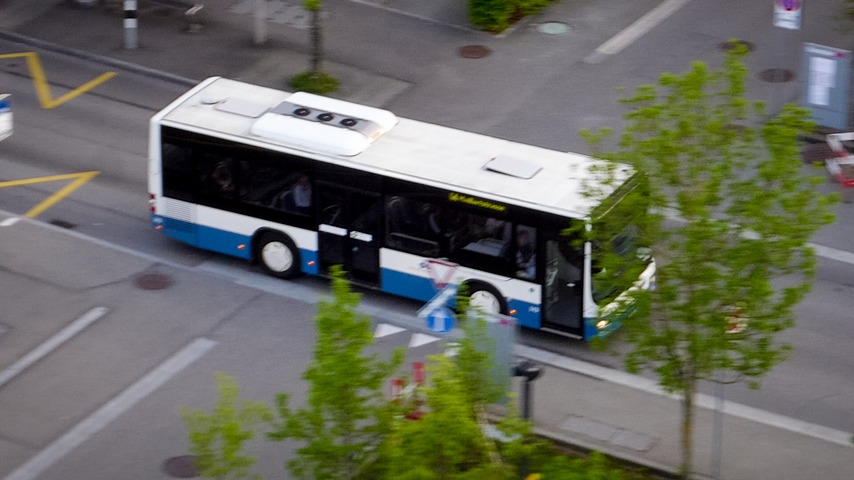} \\
\includegraphics[width=0.32\linewidth]{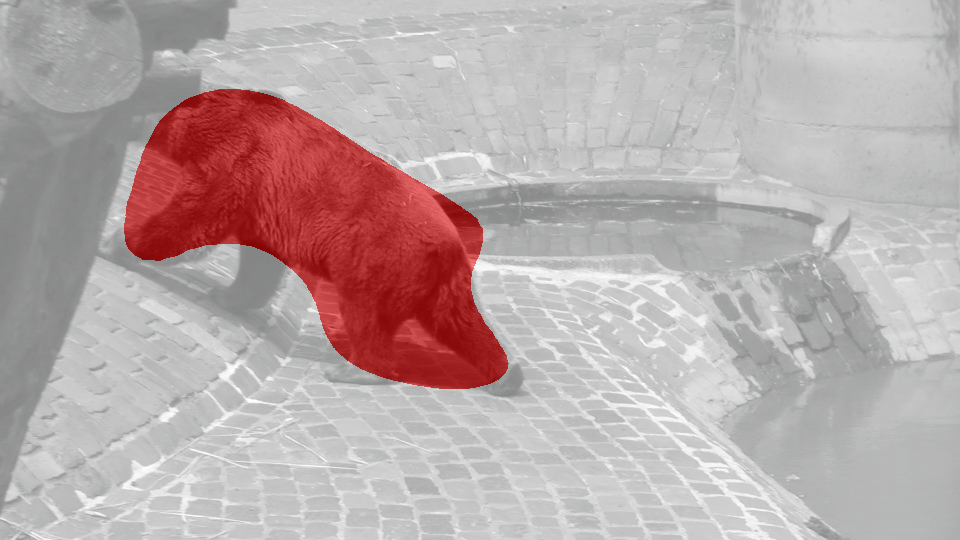}
\includegraphics[width=0.32\linewidth]{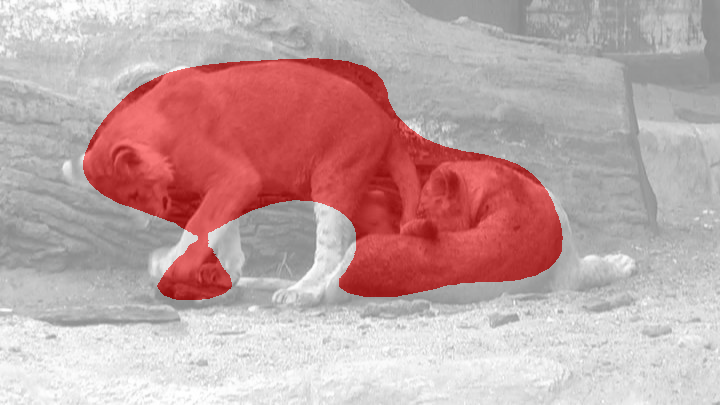}
\includegraphics[width=0.32\linewidth]{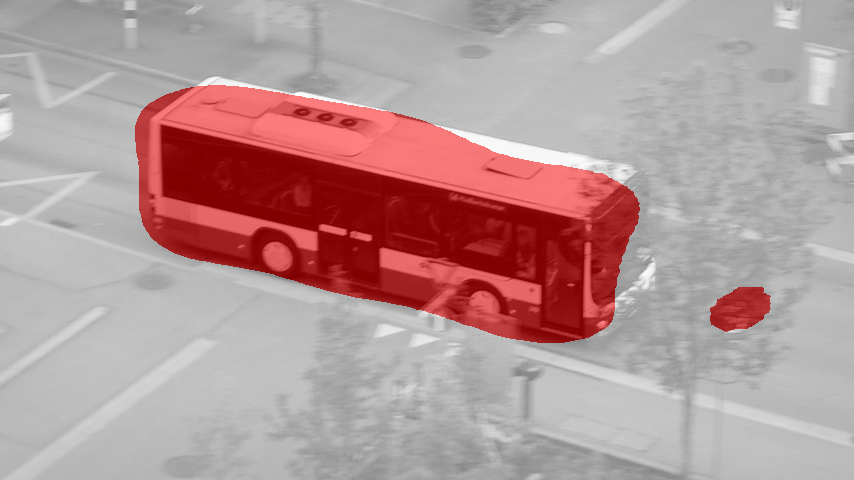} \\
\vfill
\includegraphics[height=0.197\linewidth]{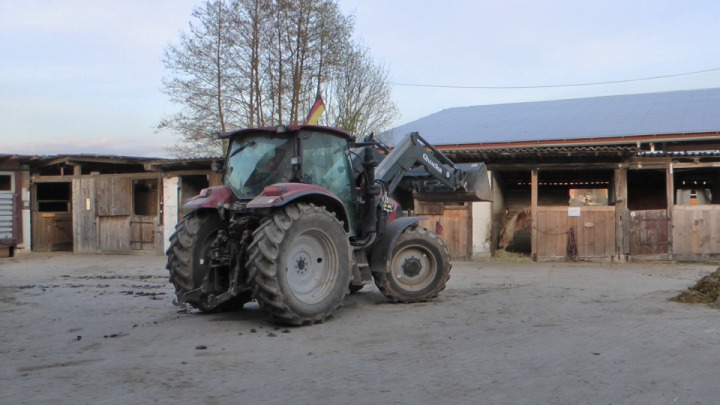}
\includegraphics[height=0.197\linewidth]{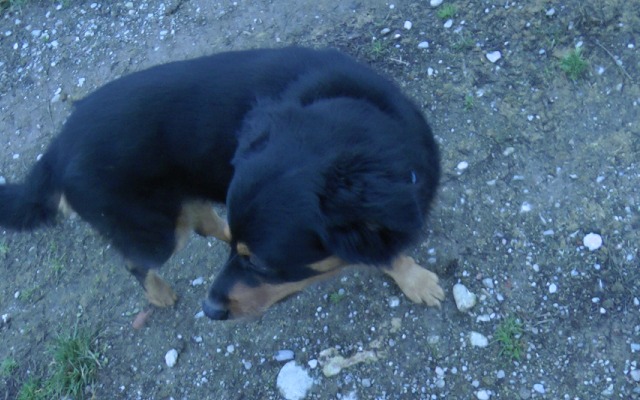}
\includegraphics[height=0.197\linewidth]{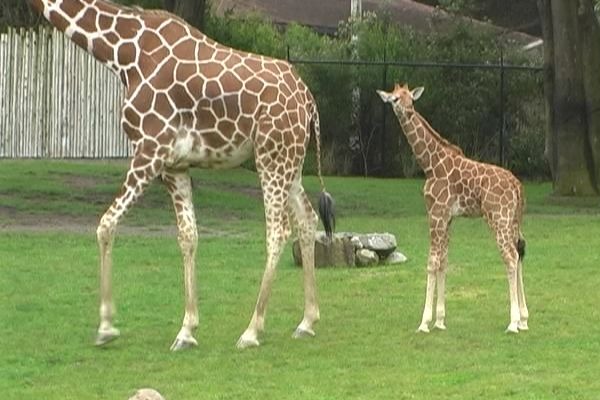} \\
\includegraphics[height=0.197\linewidth]{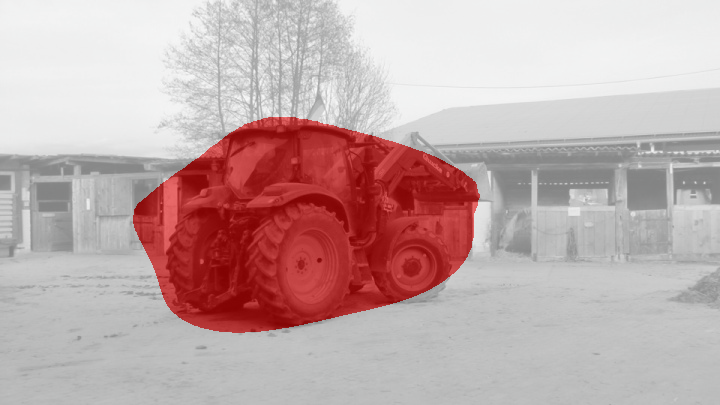}
\includegraphics[height=0.197\linewidth]{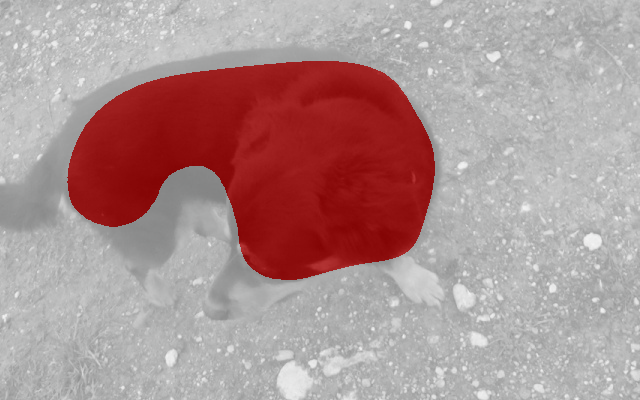}
\includegraphics[height=0.197\linewidth]{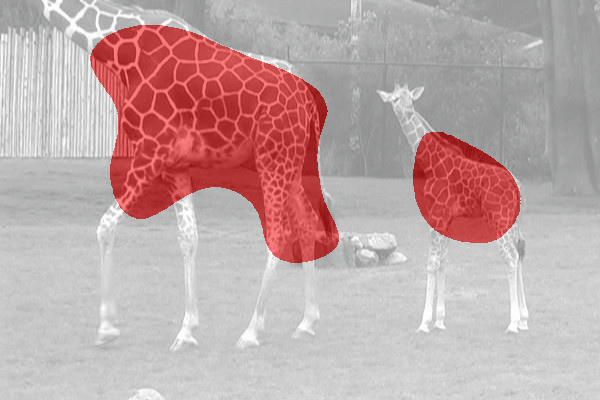}
\caption{Examples of segmentations produced by our ConvNet on held out images. The ConvNet is able to identify the motile object (or objects) and segment it out from a single frame. Masks are not perfect but they do capture the general object shape.}\vspace{-1mm}
\label{fig:davis_figs}
\end{figure}

We find that uNLC often fails on videos in the wild.
Sometimes this is because the assumption of there being a single moving object in the video is not satisfied, especially in long videos made up of multiple shots showing different objects.
We use a publicly available appearance-based shot detection method~\cite{potapov2014category} (also unsupervised) to divide the video into shots and run uNLC separately on each shot.

Videos in the wild are also often low resolution and have compression artifacts, which can degrade the resulting segmentations.
From our experiments using strong supervision, we know our approach can be robust to such noise.
Nevertheless, since a large video dataset comprises a massive collection of frames, we simply discard badly segmented frames based on two heuristics.
Specifically, we discard: (1) frames with too many ($>$80\%) or too few ($<$10\%) pixels marked as foreground; (2) frames with too many pixels ($>$10\%) within 5\% of the frame border that are marked as foreground.
In preliminary tests, we found that results were not sensitive to the precise thresholds used.

We ran uNLC on videos from YFCC100m~\cite{thomee2016yfcc100m}, which contains about 700,000 videos.
After pruning, we ended up with 205,000 videos.
We sampled 5-10 frames per shot from each video to create our dataset of 1.6M images, so we have slightly more frames than images in ImageNet.
However, note that our frames come from fewer videos and are therefore more correlated than images from ImageNet.

We stress that our approach in generating this dataset is completely unsupervised, and does not use any form of supervised learning in any part of the pipeline.
The code for the segmentation and pruning, together with our automatically generated dataset of frames and segments, will be made publicly available soon.

Our motion segmentation approach is far from state-of-the-art, as can be seen by the noisy segments shown in Figure~\ref{fig:motionsegegs}.
Nevertheless, we find that our representation is quite resilient to this noise (as shown below). As such, we did not aim to improve the particulars of our motion segmentation.

\subsection{Learning to Segment from Noisy Labels}
As before, we feed the ConvNet cropped images, jittered in scale and translation, and ask it to predict the motile foreground object.
Since the motion segmentation output is noisy, we do not trust the absolute foreground probabilities it provides.
Instead, we convert it into a \emph{trimap} representation in which pixels with a probability $<$0.4 are marked as negative samples, those with a probability $>$0.7 are marked as positives, and the remaining pixels are marked as ``don't cares'' (in preliminary experiments, our results were found to be robust to these thresholds).
The ConvNet is trained with a logistic loss only on the positive and negative pixels; don't care pixels are ignored.
Similar techniques have been successfully explored earlier in segmentation~\cite{arteta2016counting,kohli2009robust}.

Despite the steps we take to get good segments, the uNLC output is still noisy and often grossly incorrect, as can be seen from the second column of Figure~\ref{fig:motionsegegs}.
However, if there are no \emph{systematic} errors, then these motion-based segments can be seen as perturbations about a true latent segmentation.
Because a ConvNet has finite capacity, it will not be able to fit the noise perfectly and might instead learn something closer to the underlying correct segmentation.

Some positive evidence for this can be seen in the output of the trained ConvNet on its training images (Fig.~\ref{fig:motionsegegs}, third column).
The ConvNet correctly identifies the motile object and its rough shape, leading to a smoother, more correct segmentation than the original motion segmentation.

The ConvNet is also able to \emph{generalize} to unseen images.
Figure~\ref{fig:davis_figs} shows the output of the ConvNet on frames from the DAVIS~\cite{Perazzi2016}, FBMS~\cite{fbms} and VSB~\cite{vbs} datasets, which were not used in training.
Again, it is able to identify the moving object and its rough shape from just a single frame.
When evaluated against human annotated segments in these datasets, we find that the ConvNet's output is significantly \emph{better} than the uNLC segmentation output
as shown below:
\begin{center}
\footnotesize{
\begin{tabular}{lcc}
Metric & uNLC & ConvNet (unsupervised) \\
\midrule
Mean IoU (\%) & 13.1 & \textbf{24.8} \\
Precision (\%) & 15.4 & \textbf{29.9} \\
Recall (\%) & 45.8 & \textbf{59.3} \\
\end{tabular}
}
\end{center}

\begin{table*}
\resizebox{\linewidth}{!}{
\addtolength{\tabcolsep}{2pt}
\begin{tabular}{lrrrrrrcrrrrrrrc}\\
 & \multicolumn{6}{c}{\bf Full train set} &
 & \multicolumn{6}{c}{\bf 150 image set} & \\
Method & All & $>$c1 & $>$c2 & $>$c3 & $>$c4 & $>$c5 &
       & All & $>$c1 & $>$c2 & $>$c3 & $>$c4 & $>$c5 && \#wins \\
\toprule
\emph{Supervised}\\
~~Imagenet
  & 56.5 & 57.0 & 57.1 & 57.1 & 55.6 & 52.5 && 17.7 & 19.1 & 19.7 & 20.3 & 20.9 & 19.6 && NA\\
~~Sup. Masks (Ours)
  & 51.7 & 51.8 & 52.7 & 52.2 & 52.0 & 47.5 && 13.6 & 13.8 & 15.5 & 17.6 & 18.1 & 15.1 && NA\\
\midrule
\emph{Unsupervised}\\
~~Jigsaw$^\ddagger$~\cite{NorooziECCV2016}
  & 49.0 & 50.0 & 48.9 & 47.7 & 45.8 & 37.1 && 5.9 & 8.7 & 8.8 & 10.1 & 9.9 & 7.9 && NA\\
~~Kmeans~\cite{krahenbuhl2015data}
  & 42.8 & 42.2 & 40.3 & 37.1 & 32.4 & 26.0 && 4.1 & 4.9 & 5.0 & 4.5 & 4.2 & 4.0 && 0\\
~~Egomotion~\cite{agrawal2015learning}
  & 37.4 & 36.9 & 34.4 & 28.9 & 24.1 & 17.1 && -- & -- & -- & -- & -- & -- && 0\\
~~Inpainting~\cite{pathakCVPR16context}
  & 39.1 & 36.4 & 34.1 & 29.4 & 24.8 & 13.4 && -- & -- & -- & -- & -- & -- && 0\\
~~Tracking-gray~\cite{wang2015unsupervised}
  & 43.5 & 44.6 & 44.6 & 44.2 & 41.5 & 35.7 && 3.7 & 5.7 & 7.4 & 9.0 & 9.4 & 9.0 && 0\\
~~Sounds~\cite{owens2016ambient}
  & 42.9 & 42.3 & 40.6 & 37.1 & 32.0 & 26.5 && 5.4 & 5.1 & 5.0 & 4.8 & 4.0 & 3.5 && 0\\
~~BiGAN~\cite{donahue2016adversarial}
  & 44.9 & 44.6 & 44.7 & 42.4 & 38.4 & 29.4 && 4.9 & 6.1 & 7.3 & 7.6 & 7.1 & 4.6 && 0\\
~~Colorization~\cite{zhang2016colorful}
  & 44.5 & 44.9 & 44.7 & 44.4 & 42.6 & 38.0 && 6.1 & 7.9 & 8.6 & 10.6 & 10.7 & 9.9 && 0\\
~~Split-Brain Auto~\cite{zhang2016split}
  & 43.8 & 45.6 & 45.6 & 46.1 & 44.1 & 37.6 && 3.5 & 7.9 & 9.6 & 10.2 & 11.0 & 10.0 && 0\\
~~Context~\cite{doersch2015unsupervised}
  & \bf49.9 & \textbf{48.8} & 44.4 & 44.3 & 42.1 & 33.2 && 6.7 & \bf10.2 & 9.2 & 9.5 & 9.4 & 8.7 && 3\\
~~Context-videos$^\dagger$~\cite{doersch2015unsupervised}
  & 47.8 & 47.9 & 46.6 & \bf47.2 & 44.3 & 33.4 && 6.6 & 9.2 & 10.7 & 12.2 & 11.2 & 9.0 && 1\\
~~Motion Masks (Ours)
  & 48.6 & 48.2 & \bf48.3 & 47.0 & \bf45.8 & \bf40.3 && \bf10.2 & \bf10.2 & \bf11.7 & \bf12.5 & \bf13.3 & \bf11.0 && \bf9\\
\end{tabular}}\\
\caption{Object detection AP (\%) on PASCAL VOC 2012 using Fast R-CNN with various pretrained ConvNets. All models are trained on \texttt{train} and tested on \texttt{val} using consistent Fast R-CNN settings. `--' means training didn't converge due to insufficient data. Our approach achieves the best performance in the majority of settings.
$^\dagger$Doersch~\etal~\cite{doersch2015unsupervised} trained their original context model using ImageNet images. The Context-videos model is obtained by retraining their approach on our video frames from YFCC. This experiment controls for the effect of the distribution of training images and shows that the image domain used for training does not significantly impact performance.
$^\ddagger$Noroozi~\etal~\cite{NorooziECCV2016} use a more computationally intensive ConvNet architecture ($>$2$\times$ longer to finetune) with a finer stride at conv1, preventing apples-to-apples comparisons. Nevertheless, their model works significantly worse than our representation when either layers are frozen or in case of limited data and is comparable to ours when network is finetuned with full training data.}
\label{tab:det}
\end{table*}

\begin{figure*}
\centering
\subfigure[Performance vs. Finetuning]{\label{fig:oursvsothers}\includegraphics[width=0.64\textwidth]{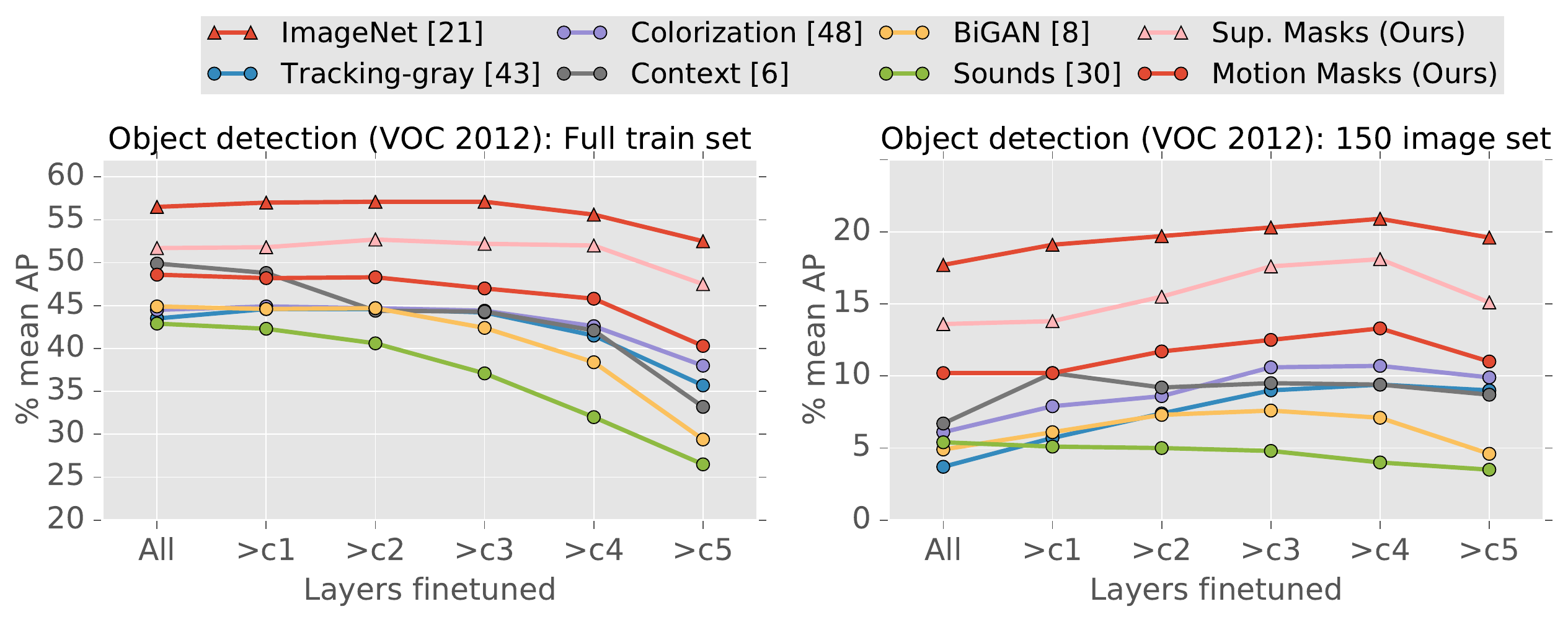}}
\quad
\subfigure[Performance vs. Data]{\label{fig:videodata}\includegraphics[width=0.32\textwidth]{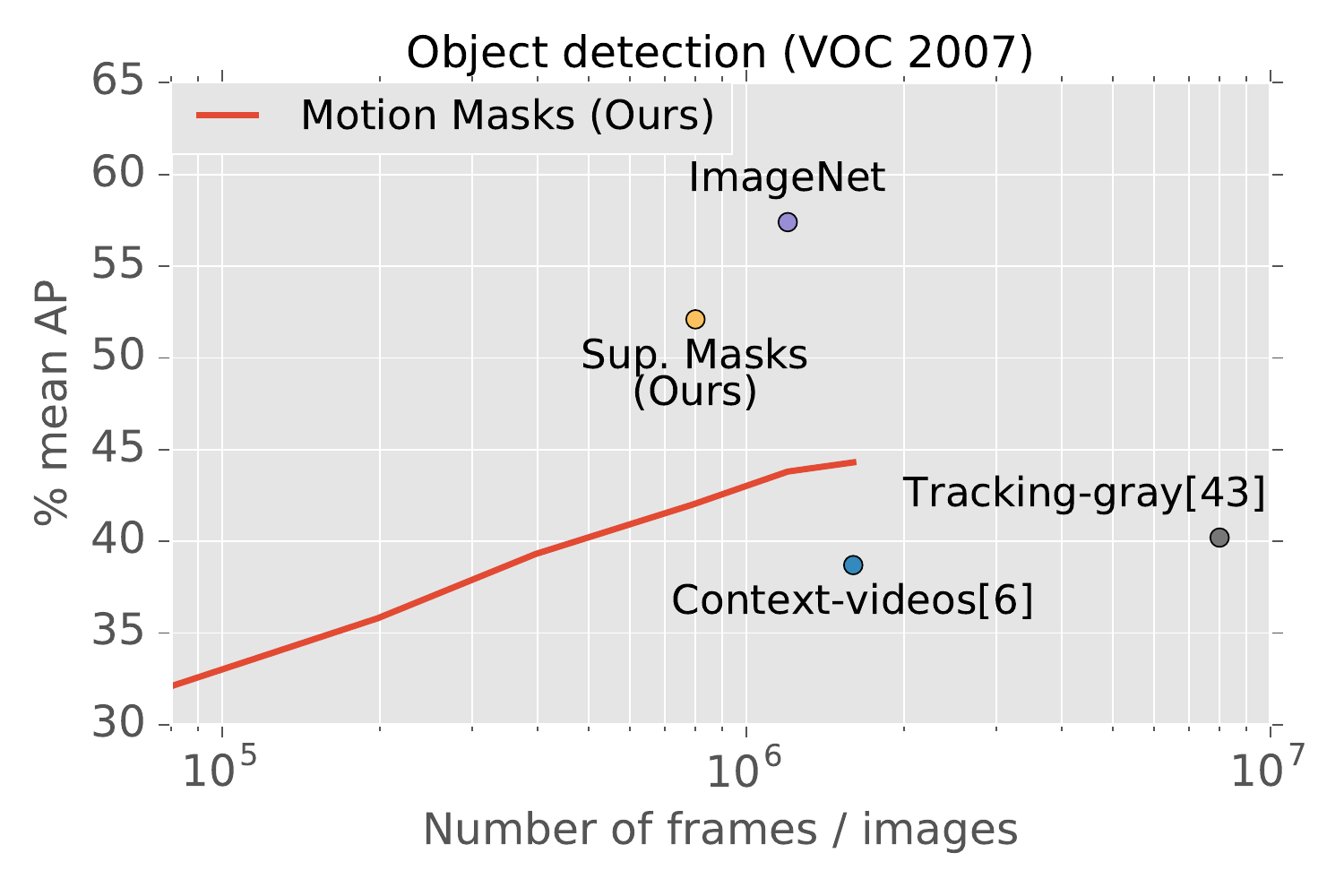}}
\caption{Results on object detection using Fast R-CNN. (a) VOC 2012 object detection results when the ConvNet representation is frozen to different extents. We compare to other unsupervised and supervised approaches. \textbf{Left}: using the full training set. \textbf{Right}: using only 150 training images (note the different y-axis scales). (b) Variation of representation quality (mean AP on VOC 2007 object detection with conv5 and below frozen) with number of training frames. A few other methods are also shown. Context-videos~\cite{doersch2015unsupervised} is the representation of Doersch \etal~\cite{doersch2015unsupervised} retrained on our video frames. Note that most other methods in Table~\ref{tab:det} use ImageNet as their train set.}
\label{fig:detresults_combined}
\end{figure*}

These results confirm our earlier finding that the ConvNet is able to learn well even from noisy and often incorrect ground truth.
However, the goal of this paper is not segmentation, but representation learning.
We evaluate the learned representation in the next section.

\section{Evaluating the Learned Representation}
\subsection{Transfer to Object Detection}
We first evaluate our representation on the task of object detection using Fast R-CNN.
We use VOC 2007 for cross-validation: we pick an appropriate learning rate for each method out of a set of 3 values  $\{0.001, 0.002 \textrm{ and } 0.003\}$.
Finally, we train on VOC 2012 train and test on VOC 2012 val exactly once.
We use multi-scale training and testing and discard difficult objects during training.

We present results with the ConvNet parameters frozen to different extents.
As discussed in Section~\ref{sec:evaluationprotocol}, a good representation should work well both as an initialization to fine-tuning and also when most of the ConvNet is frozen.

We compare our approach to ConvNet representations produced by recent prior work on unsupervised learning~\cite{agrawal2015learning, doersch2015unsupervised, donahue2016adversarial, NorooziECCV2016, owens2016ambient, pathakCVPR16context, wang2015unsupervised, zhang2016colorful}.
We use publicly available models for all methods shown.
Like our ConvNet representation, all models have the AlexNet architecture, but differ in minor details such as the presence of batch normalization layers~\cite{doersch2015unsupervised} or the presence of grouped convolutions~\cite{zhang2016colorful}.

We also compare to two models trained with strong supervision.
The first is trained on ImageNet classification.
The second is trained on manually-annotated segments (without class labels) from COCO (see Section~\ref{sec:super}).

Results are shown in Figure~\ref{fig:oursvsothers} (left) and Table~\ref{tab:det} (left).
We find that our representation learned from unsupervised motion segmentation performs on par or better than prior work on unsupervised learning across all scenarios.

As we saw in Section~\ref{sec:control}, in contrast to ImageNet supervised representations, the representations learned by previous unsupervised approaches show a large decay in performance as more layers are frozen, owing to the representation becoming highly specific to the pretext task.
Similar to our \emph{supervised} approach trained on segmentations from COCO, we find that our \emph{unsupervised} approach trained on motion segmentation also shows \emph{stable} performance as the layers are frozen.
Thus,  unlike prior work on unsupervised learning, the upper layers in our representation learn high-level abstract concepts that are useful for recognition.

It is possible that some of the differences between our method and prior work are because the training data is from different domains (YFCC100m videos \vs ImageNet images).
To control for this, we retrained the model from~\cite{doersch2015unsupervised} on frames from our video dataset (see Context-videos in Table~\ref{tab:det}).
The two variants perform similarly: 33.4\% mean AP when trained on YFCC with conv5 and below frozen compared to 33.2\% for the ImageNet version.
This confirms that the different image sources do not explain our gains.

\subsection{Low-shot Transfer}
A good representation should also aid learning when training data is scarce, as we motivated in Section~\ref{sec:evaluationprotocol}.
Figure~\ref{fig:oursvsothers} (right) and Table~\ref{tab:det} (right) show how we compare to other unsupervised and supervised approaches on the task of object detection when we have few (150) training images.
We observe that in this scenario it actually hurts to fine-tune the entire network, and the best setup is to leave some layers frozen.
Our approach provides the best AP overall (achieved by freezing all layers up to and including conv4) among all other representations from recent unsupervised learning methods by a large margin.
The performance in other low-shot settings is presented in Figure~\ref{fig:oursvsothersall}.

Note that in spite of its strong performance relative to prior unsupervised approaches, our representation learned without supervision on video trails both the strongly supervised mask and ImageNet versions by a significant margin. We discuss this in the following subsection.

\begin{figure*}
\centering
\includegraphics[width=0.99\linewidth]{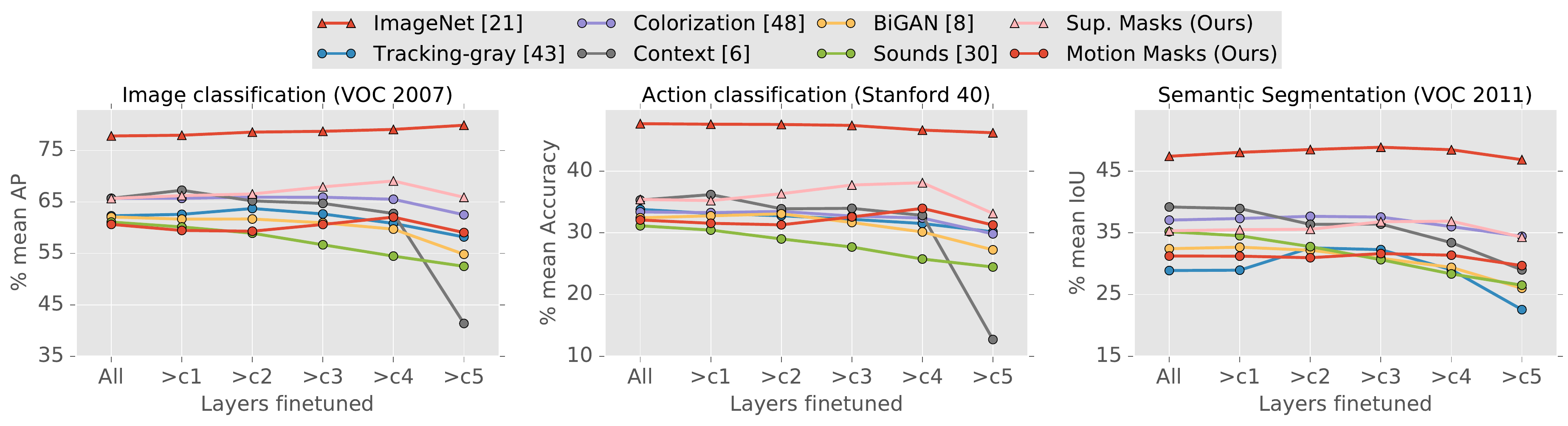}
\caption{Results on image (object) classification on VOC 2007, single-image action classification on Stanford 40 Actions, and semantic segmentation on VOC 2011. Results shown with ConvNet layers frozen to different extents (note that the metrics vary for each task).}
\vspace{-3mm}
\label{fig:oursvsothersclass}
\end{figure*}

\subsection{Impact of Amount of Training Data}
The quality of our representation (measured by Fast R-CNN performance on VOC 2007 with all conv layers frozen) grows roughly logarithmically with the number of frames used.
With 396K frames (50K videos), it is already better than prior state-of-the-art~\cite{doersch2015unsupervised} trained on a million ImageNet images, see Figure~\ref{fig:videodata}.
With our full dataset (1.6M frames) accuracy increases substantially.
If this logarithmic growth continues, our representation will be on par with one trained on ImageNet if we use about 27M frames (or 3 to 5 million videos, the same order of magnitude as the number of images in ImageNet).
Note that frames from the same video are very correlated.
We expect this number could be reduced with more algorithmic improvements.

\subsection{Transfer to Other Tasks}
As discussed in Section~\ref{sec:evaluationprotocol}, a good representation should generalize across tasks.
We now show experiments for two other tasks: image classification and semantic image segmentation.
For image classification, we test on both object and action classification.

\paragraph{Image Classification.}
We experimented with image classification on PASCAL VOC 2007 (object categories) and Stanford 40 Actions~\cite{yao2011human} (action labels).
To allow comparisons to prior work~\cite{donahue2016adversarial,zhang2016colorful}, we used random crops during training and averaged scores from 10 crops during testing (see~\cite{donahue2016adversarial} for details).
We minimally tuned some hyper-parameters (we increased the step size to allow longer training) on VOC 2007 validation, and used the same settings for both VOC 2007 and Stanford 40 Actions.
On both datasets, we trained with different amounts of fine-tuning as before.
Results are in the first two plots in Figure~\ref{fig:oursvsothersclass}.

\paragraph{Semantic Segmentation.}
We use fully convolutional networks for semantic segmentation with the default hyper-parameters~\cite{long2014fully}.
All the pretrained ConvNet models are finetuned on union of images from VOC 2011 train set and additional SBD train set released by Hariharan~\etal~\cite{hariharan2011semantic}, and we test on the VOC 2011 val set after removing overlapping images from SBD train.
The last plot in Figure~\ref{fig:oursvsothersclass} shows the performance of different methods when the number of layers being finetuned is varied.

\paragraph{Analysis.}
Like object detection, all these tasks require semantic knowledge.
However, while in object detection the ConvNet is given a tight crop around the target object, the input in these image classification tasks is the entire image, and semantic segmentation involves running the ConvNet in a sliding window over all locations.
This difference appears to play a major role.
Our representation was trained on object crops, which is similar to the setup for object detection, but quite different from the setups in Figure~\ref{fig:oursvsothersclass}.
This mismatch may negatively impact the performance of our representation, both for the version trained on motion segmentation and the strongly supervised version.
Such a mismatch may also explain the low performance of the representation trained by Wang \etal~\cite{wang2015unsupervised} on semantic segmentation.

Nevertheless, when the ConvNet is progressively frozen, our approach is a strong performer.
When all layers until conv5 are frozen, our representation is better than other approaches on action classification and second only to colorization~\cite{zhang2016colorful} on image classification on VOC 2007 and semantic segmentation on VOC 2011.
Our higher performance on action classification might be due to the fact that our video dataset has many people doing various actions.

\section{Discussion}
We have presented a simple and intuitive approach to unsupervised learning by using segments from low-level motion-based grouping to train ConvNets.
Our experiments show that our approach enables effective transfer especially when computational or data constraints limit the amount of task-specific tuning we can do.
Scaling to larger video datasets should allow for further improvements.

We noted in Figure~\ref{fig:motionsegegs} that our network learns to refine the noisy input segments.
This is a good example of a scenario where ConvNets can learn to extract signal from large amounts of noisy data.
Combining the refined, single-frame output from the ConvNet with noisy motion cues extracted from the video should lead to better pseudo ground truth, and can be used by the ConvNet to bootstrap itself.
We leave this direction for future work.

\begin{figure*}
\centering
\includegraphics[width=0.33\linewidth]{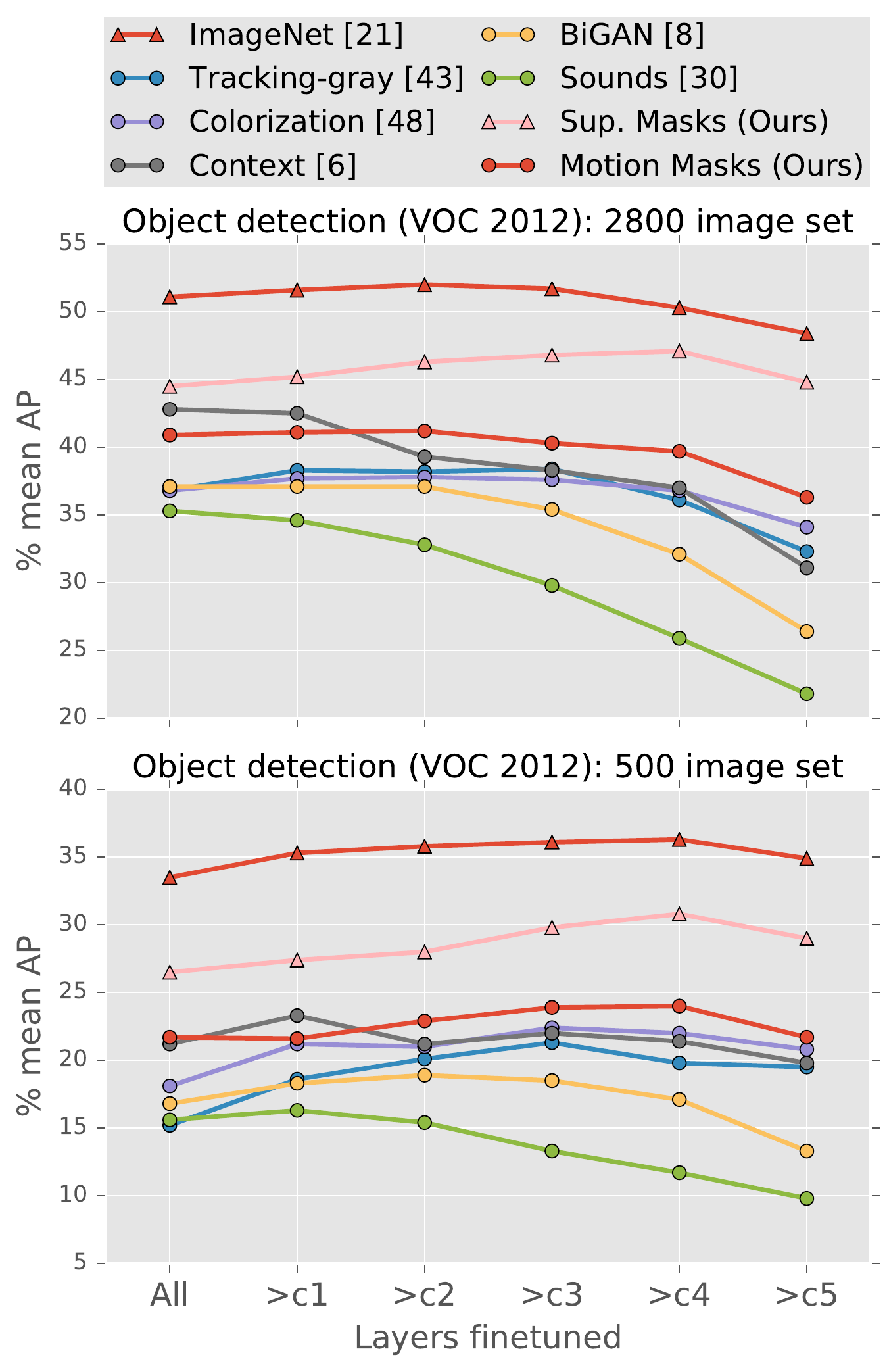}
\includegraphics[width=0.33\linewidth]{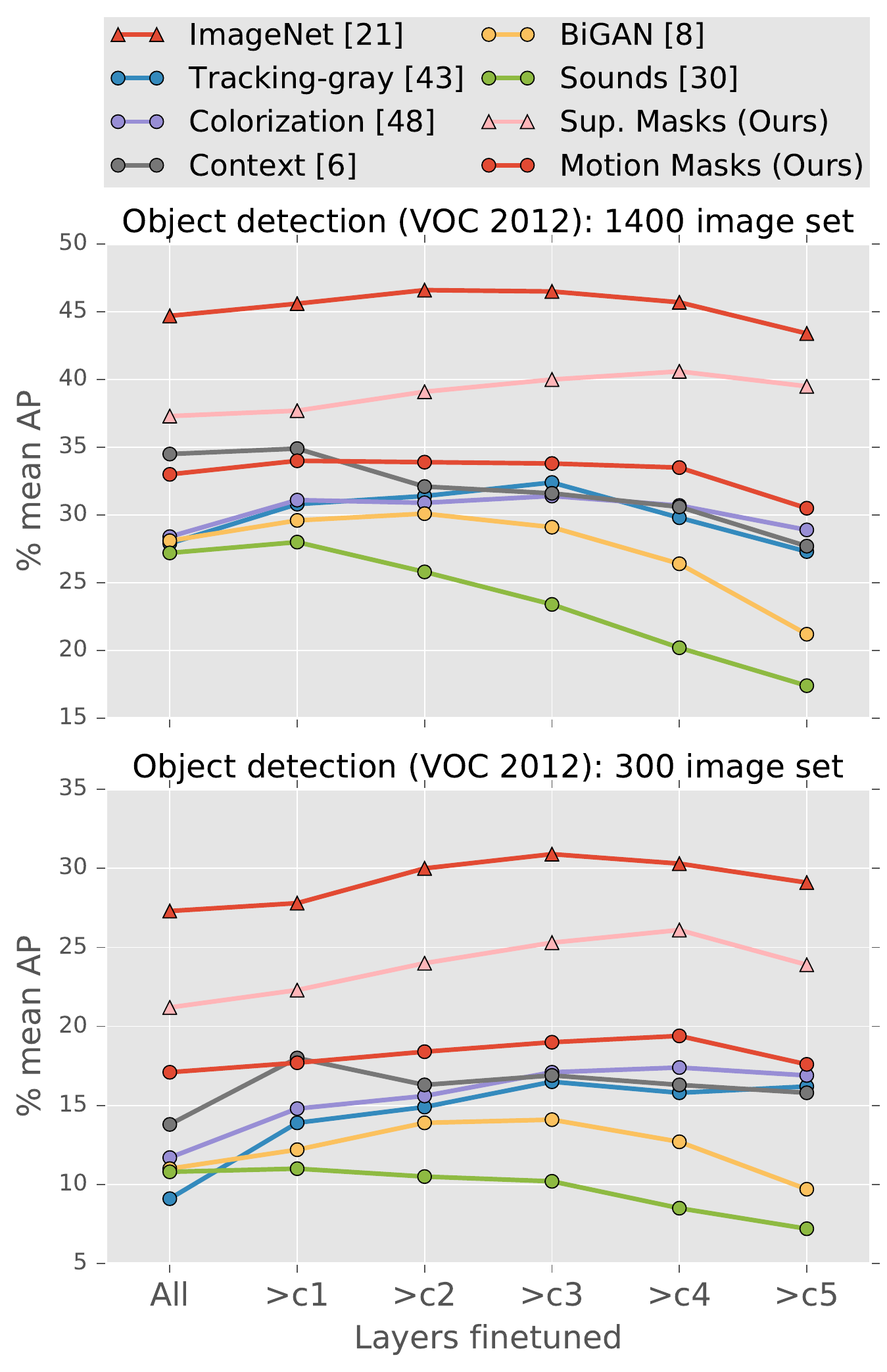}
\includegraphics[width=0.33\linewidth]{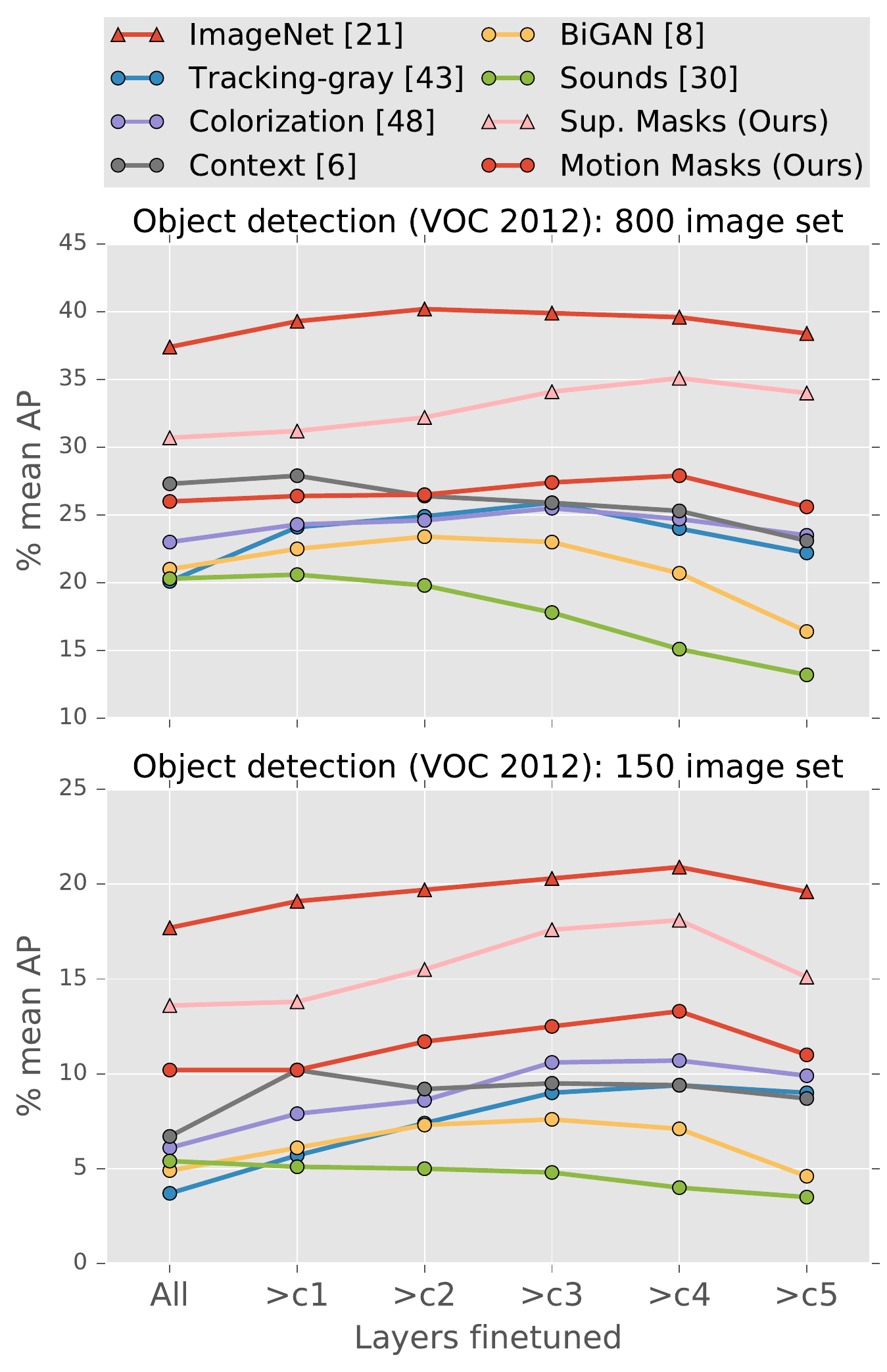}
\caption{Results for object detection on Pascal VOC 2012 using Fast R-CNN and varying number of images available for finetuning. Each plot shows the comparison of different unsupervised learning methods as the number of layers being finetuned is varied. Different plots depict this variation for different amounts of data available for finetuning Fast R-CNN (please note the different y-axis scales for each plot). As the data for finetuning decreases, it is actually better to freeze more layers. Our method works well across all the settings and scales and as the amount of data decreases. When layers are frozen or data is limited, our method significantly outperforms other methods. This suggests that features learned in the higher layers of our model are good for recognition.}
\label{fig:oursvsothersall}
\end{figure*}

{\small
\bibliographystyle{ieee}
\bibliography{main}
}

\end{document}